%% file: preprint.tex
\documentclass{article} % For LaTeX2e
\usepackage{iclr2027_conference,times}

\input{math_commands.tex}

\usepackage{hyperref}
\usepackage{url}
\usepackage{xcolor}

\usepackage{booktabs}
\usepackage{multirow}
\usepackage{graphicx}
\usepackage{enumitem}
\usepackage{caption}
\usepackage{capt-of}
\usepackage{wrapfig}
\usepackage{pifont}
\usepackage{colortbl}
\usepackage{fontawesome5}

\usepackage{tikz}

\definecolor{GammaPurple}{HTML}{601886}
\colorlet{GammaCite}{GammaPurple!80!white}
\colorlet{GammaFrame}{GammaPurple!60!white}
\colorlet{GammaBack}{GammaPurple!5!white}

\providecommand{\ptYes}{\ding{51}}
\providecommand{\ptNo}{\ding{55}}
\definecolor{policyrow}{RGB}{237,234,255}

\hypersetup{colorlinks=true,citecolor=GammaCite,linkcolor=GammaPurple,urlcolor=GammaPurple}

\title{V-JEPA Policy: Building Effective World-Action Models on Predictive Visual Latents}

\author{
Yang Zhang$^{1}$,
Jiangyuan Zhao$^{2}$,
Chenyou Fan$^{3}$,
Jiayu Hu$^{4}$,
Xiu Yuan$^{5}$,
Chenjia Bai$^{6,7}$,\\
\,\,\textbf{Xiu Li}$^{1}$\thanks{Corresponding author.}\\ 
\textsuperscript{1}Tsinghua University, \textsuperscript{2}Shanghai Jiao Tong University, \textsuperscript{3}Fudan University,\\
\textsuperscript{4}University of Science and Technology of China,
\textsuperscript{5}Washington University in St. Louis,\\
\textsuperscript{6}The Institute of Artificial Intelligence, China Telecom (TeleAI),
\textsuperscript{7}Gamma Robotics ($\gamma$)\\
\texttt{~z-yang21@mails.tsinghua.edu.cn, li.xiu@sz.tsinghua.edu.cn}\\[1em]
\makebox[\linewidth][c]{%
  \begin{tabular}{@{}l@{}}
    Code: \href{https://github.com/breez3young/VJEPA-Policy}%
      {https://github.com/breez3young/VJEPA-Policy}\\[3pt]
    Project Page: \href{https://breez3young.github.io/vjepa-policy/}%
      {https://breez3young.github.io/vjepa-policy/}
  \end{tabular}%
}
\vspace{-1em}
}

\iclrfinalcopy % Uncomment for camera-ready version, but NOT for submission.
\begin{document}

\maketitle

\input{secs/abstract}
\input{secs/introduction}
\input{secs/preliminary}
\input{secs/method}
\input{secs/related_work}
\input{secs/experiments}

\vspace{-1.2em}
\section{Conclusion}\label{sec:conclusion}
\vspace{-0.6em}
We showed that effective WAM learning can build on the latent space of a frozen predictive visual encoder without inheriting a complete pretrained visual generator.
V-JEPA Policy instantiates this approach by jointly training a future predictor and an action expert from scratch, coupled through future-informed context states.
The resulting compact policy achieves competitive simulation performance and supports real-world bimanual manipulation.
Controlled encoder comparisons favor predictive visual latents, particularly under distribution shifts.
Predictor-only pretraining on DROID video--instruction pairs without using action labels further improves downstream control and robustness, with LIBERO-Plus gains that extended scratch training does not recover within the evaluated budget.
Together, these findings support predictive visual latents as a sufficient foundation for learning WAMs directly from task-specific demonstrations and acquiring transferable future-modeling knowledge from broader video experience.

% \section*{Acknowledgments}
% We would like to thank Jianing Ye for his insightful discussions and comments.

% \section*{AI use statement}
% In this work, we used generative AI tools to polish and edit the paper for readability and to create preliminary design sketches for figures. We have reviewed all AI-assisted work, checked and manually revised the generated content, and drawn the final figures ourselves. We take responsibility for the final content of this work.

% \section*{Ethics statement}
% We have read and adhered to the ICLR Code of Ethics. We are not aware of any ethical concerns arising from this work that require separate discussion.

% \section*{Reproducibility statement}
% Sec.~\ref{sec:method} describes the model architecture and training objectives. Sec.~\ref{sec:experiments} and the appendix document the experiment settings, training details, and evaluation protocols. The publicly available datasets and pretrained encoders used in this work are cited in the paper.

\bibliography{iclr2027_conference}
\bibliographystyle{iclr2027_conference}

\newpage
\appendix
\input{secs/appendix}

\end{document}

%% file: math_commands.tex
\usepackage{amsmath,amsfonts,bm}

\def\eqref#1{equation~\ref{#1}}
\def\1{\bm{1}}

\DeclareMathAlphabet{\mathsfit}{\encodingdefault}{\sfdefault}{m}{sl}
\SetMathAlphabet{\mathsfit}{bold}{\encodingdefault}{\sfdefault}{bx}{n}

%% file: secs/abstract.tex
\begin{abstract}
World-action models (WAMs) couple future visual-state prediction with action generation.
By adapting video generators or image-editing models pretrained at scale, a prominent line of recent WAMs inherits both predictive knowledge and the models in which it was learned.
We ask whether a predictive visual latent space induced by large-scale predictive pretraining can instead provide a sufficient foundation for effective WAM learning without inheriting a complete pretrained visual generative model.
To answer this question, we introduce \textbf{V-JEPA Policy}, a simple framework that builds a WAM on the latent space of a frozen V-JEPA 2.1 encoder.
An instruction-conditioned future-latent predictor and a flow-matching action expert are jointly learned from scratch in a single downstream stage, with the predictor's future-informed context key--value states conditioning action generation.
With 0.9B total parameters, of which 0.6B are trainable, \textbf{V-JEPA Policy} achieves competitive performance with representative WAM and vision-language-action baselines across LIBERO, LIBERO-Plus, and RoboCasa-GR1. Comparing visual foundations under the same downstream framework and training budget identifies V-JEPA latents as more effective than the discriminative, reconstructive, and video-understanding-oriented alternatives, particularly under distribution shifts.
Beyond task-specific learning, the same latent space supports acquiring transferable future-modeling knowledge from diverse in-the-wild instruction-annotated videos.
Pretraining the predictor on DROID video--instruction pairs without action labels and adapting it into a WAM within our framework yields substantial gains in downstream control and out-of-distribution generalization.
Together, these findings establish predictive visual latents as a foundation for learning effective WAMs, supporting both direct learning from task-specific demonstrations and the transfer of future-modeling knowledge acquired through predictor pretraining on broader in-the-wild videos.
% Our code is available at \url{https://github.com/breez3young/VJEPA-Policy}.
% Together, these findings establish a route to learning effective WAMs through predictive visual representations, supporting both direct learning from task-specific demonstrations, and transfer of future-modeling knowledge acquired through predictor pretraining on broad in-the-wild video data.

\end{abstract}

%% file: secs/introduction.tex
\section{Introduction}\label{sec:intro}
\vspace{-0.6em}
Advanced physical intelligence should endow robots with the capability not only to understand the current scene but also to anticipate how it may change through interaction. 
World-action models (WAMs) have thus emerged as a promising paradigm for instantiating this principle by coupling action generation with future visual-state prediction, demonstrating strong task performance and improved generalization to unfamiliar tasks and environments \citep{zhu2025unifiedworldmodels, pai2025mimicvideos, liang2025videopolicy, kim2026cosmospolicy, ye2026dreamzero, bi2026motus, li2026lingbot-va, zhang2026lingbot-va2}.
They highlight the value of predictive knowledge for general-purpose robot control.

A prominent line of work acquires this capability by adapting visual generative models pretrained at scale for effective WAM learning. Video-generation models provide priors over motion and scene evolution learned from large-scale video data \citep{pai2025mimicvideos, kim2026cosmospolicy, yuan2026fastwam}, while image-editing models provide priors over instruction-conditioned visual transformations, mapping a current observation directly to a task-specified target visual state without generating intermediate frames \citep{bfl2025flux2, wu2025qwen-image, zhang2026imagewam}. Despite these different formulations, both approaches transfer knowledge about how visual states change together with the generative backbone in which that knowledge was learned.
This raises a natural question:
\begin{center}
\emph{Does effective WAM learning require inheriting a complete pretrained visual generative model, or can a predictive visual latent space learned at scale provide a sufficient foundation?}
\end{center}

To answer this question, we introduce \textbf{V-JEPA Policy}, a framework that builds a WAM on the predictive latent space of a frozen V-JEPA 2.1 encoder~\citep{bardes2024vjepa, assran2025vjepa2, mur2026vjepa2_1}.
We jointly learn an instruction-conditioned future-latent predictor and a flow-matching action expert directly from task-specific demonstrations.
The predictor follows the vision transformer architecture of the V-JEPA predictor, augmented with cross-attention for instruction conditioning, and combines observed latent tokens with learnable queries at future positions to forecast future visual latents in a single forward pass.
Inspired by recent foundation policies~\citep{black2024pi0, black2025pi05, ye2026dreamzero, zhang2026prts}, our framework adopts a Mixture-of-Transformers (MoT) architecture with shared attention to combine the future predictor and the action expert. 
Context and future queries interact through bidirectional attention, and the resulting layer-wise context key--value states condition the action expert, allowing future modeling to directly shape action generation.
V-JEPA's large-scale video predictive self-supervised pretraining provides a visual latent space suited to this construction.
With the visual encoder kept frozen and both trainable modules learned from scratch in a single downstream stage, our design directly tests whether a predictive visual latent space learned at scale can provide a sufficient foundation for effective WAM learning without inheriting a complete pretrained visual generator.

Beyond learning from task-specific demonstrations, we further investigate whether the same latent space can efficiently absorb general future-modeling knowledge from broad in-the-wild videos and transfer it to downstream control.
Task-specific demonstrations cover only a limited range of behaviors and visual transitions.
We therefore pretrain only the instruction-conditioned future-latent predictor on diverse DROID video--instruction pairs, using future-prediction supervision without action labels while keeping the V-JEPA encoder frozen~\citep{khazatsky2024droid}.
We then transfer the predictor to downstream V-JEPA Policy training, where a freshly initialized flow-matching action expert is learned under the same joint objective and downstream budget.
The latent space is thus tested not only as a foundation for direct WAM learning but also as a reliable substrate for acquiring and transferring more general future-modeling knowledge underlying in-the-wild videos.

We extensively evaluate V-JEPA Policy on both standard and distribution-shifted simulation benchmarks spanning settings from single-arm manipulation to whole-body control, as well as on a real-world dual-arm platform.
Beyond comparisons with baselines, we also systematically analyze the framework through controlled studies of its visual foundation and encoder scale, context-future interaction and future-modeling supervision, training-compute scaling, and action-free predictor pretraining and transfer.
With 0.9B total parameters, of which only 0.6B are trainable, the one-stage model reaches 97.25\% on LIBERO~\citep{liu2023libero}, 79.25\% on LIBERO-Plus~\citep{fei2025liberoplus}, 50.92\% on RoboCasa-GR1 tasks~\citep{bjorck2025gr00t, robocasa2024}, and 45\% on two real-world multi-stage bimanual coordination tasks while remaining competitive with representative WAM and VLA baselines.
Under the fixed downstream framework and training budget, the V-JEPA predictive latent outperforms the discriminative~\citep{oquab2024dinov2, simeoni2026dinov3}, reconstructive~\citep{wan2025wan}, and video-understanding-oriented alternatives~\citep{yan2026internvideo3}, with the performance gaps widening under distribution shifts.
Extending downstream training compute on the same task demonstrations yields diminishing returns in generalization.
Predictor-only pretraining on DROID video--instruction pairs without action label supervision instead raises LIBERO-Plus success rate from 79.25\% to 91.50\% at the default smaller downstream budget, remarkably surpassing the 81.64\% success rate achieved by extended training.
The same pretraining also raises the average real-world success rate from 45\% to 80\%, demonstrating the effectiveness of predictor pretraining beyond simulation.
These results show that a future-latent predictor can acquire knowledge about physical world evolution from diverse in-the-wild videos and transfer it to downstream control within the same latent space.
% Finally, predictor-only pretraining on action-free DROID video--instruction pairs raises LIBERO-Plus success rate from 79.25\% to 91.50\% at the same downstream budget, showing that a future-latent predictor can acquire knowledge about physical world evolution from diverse in-the-wild videos and transfer it to downstream control within the same latent space.

We highlight the main contributions as follows:

\begin{itemize}[leftmargin=10pt, topsep=0pt,itemsep=1pt,partopsep=1pt, parsep=1pt]
    \item We introduce \textbf{V-JEPA Policy}, a WAM built on the predictive latent space of a frozen V-JEPA encoder. By jointly learning an instruction-conditioned future-latent predictor and a flow-matching action expert from scratch in a single downstream stage, it enables effective WAM learning without inheriting a complete pretrained visual generative model.
    \item We conduct extensive cross-benchmark evaluations and controlled analyses of {V-JEPA Policy}. In particular, matched visual-substrate comparisons under a shared downstream framework and training budget show that V-JEPA's predictive visual latents yield stronger performance, with more significant gains under distribution shifts.
    \item We demonstrate that predictor-only pretraining on DROID video--instruction pairs without action supervision transfers future-modeling knowledge within the same latent space, substantially improving generalization performance on LIBERO-Plus at the same downstream budget.
\end{itemize}

%% file: secs/preliminary.tex
\vspace{-1em}
\section{Preliminaries}\label{sec:preliminary}
\vspace{-0.6em}

% 参考http://arxiv.org/abs/2601.10553 来写V-JEPA preliminary

\textbf{V-JEPA.}
The V-JEPA family of models learns visual representations through self-supervised prediction in feature space, encouraging the modeling of predictable spatiotemporal structure and dynamics rather than pixel-level
generation~\citep{lecun2022path, bardes2024vjepa, assran2025vjepa2, mur2026vjepa2_1}.
Let $y$ denote an unmasked video, $x$ a masked view of the same video, and $M$ the set of masked spatiotemporal patch positions.
A context encoder $E_{\theta}$ maps the visible patches in the masked video $x$ to context tokens, while the exponential moving average $E_{\bar\theta}$ of the context encoder computes target representations from the complete video $y$.
The predictor $P_{\phi}$ jointly processes the context tokens and learnable mask tokens $\Delta_y$ specifying the masked positions.
Both networks are vision transformers (ViTs)~\citep{dosovitskiy2020vit} with 3D rotary position embeddings (RoPE)~\citep{su2024rope}.
They are jointly trained with the latent mask-denoising objective:
{
\vspace{-10pt}
\setlength{\abovedisplayskip}{3pt}
\setlength{\belowdisplayskip}{3pt}
\setlength{\abovedisplayshortskip}{0pt}
\setlength{\belowdisplayshortskip}{0pt}
\begin{align}
\mathcal L_{\mathrm{predict}} = \frac{1}{|M|} \sum_{i\in M}
\left\| P_{\phi}\big(E_{\theta}(x),\Delta_y\big)_i - \operatorname{sg}\big(E_{\bar\theta}(y)_i\big) \right\|_1,
\label{eq:jepa}
\end{align}
}
where $\operatorname{sg}$ denotes stop-gradient and $\bar\theta$ tracks an exponential moving average of the encoder parameters $\theta$. These mechanisms are used to prevent representation collapse during joint learning.
V-JEPA 2.1 further improves dense visual representations through context-token supervision and deep self-supervision~\citep{mur2026vjepa2_1}.
We use its frozen encoder to define the visual latent space of V-JEPA Policy.
Within this space, we adapt the V-JEPA 2 predictor architecture to forecast future latents from observed context by assigning mask tokens to future positions (Sec.~\ref{sec:method}).

% %
% \textbf{Problem formulation.}
% We study language-conditioned robotic manipulation through imitation learning from a demonstration dataset $\mathcal D$. At time $t$, the robot receives RGB observations $o_t= (\mathbf{I}_t^1, ..., \mathbf{I}_t^n)$ from $n$ camera views, its proprioceptive state $\mathbf{q}_t\in\mathbb R^{d_s}$, and a task instruction $\ell$.
% Here, $o_t^{v}$ denotes the visual input from camera $v$ available at time $t$.
% Conditioned on these inputs, the policy imitates the recorded action chunk $\mathbf{A}_t=(a_t,\ldots,a_{t+H-1}) \in \mathbb{R}^{H\times d_a}$ in the demonstrations, i.e., modeling the conditional distribution $\pi (\mathbf{A}_t \mid o_t, \mathbf{q}_t, \ell)$, where $H$ is the action horizon and $d_a$ is the action dimension.
% For world-action learning, we construct training examples $(o_t,\mathbf{q}_t,\ell,\mathbf O_t^+,\mathbf A_t)$ from $\mathcal D$, where $\mathbf O_t^+$ contains visual observations sampled at future time steps of the same demonstration.
% These future observations provide supervision for latent prediction.
% At inference, the policy is conditioned on $o_t$, $\mathbf{q}_t$, and $\ell$.
% %

\textbf{Problem formulation.}
We study language-conditioned robotic manipulation through imitation learning from a demonstration dataset $\mathcal D$. At time $t$, the robot receives RGB observations $o_t=(\mathbf I_t^1,\ldots,\mathbf I_t^n)$ from $n$ camera views, its proprioceptive state $\mathbf q_t\in\mathbb R^{d_s}$, and a task instruction $\ell$.
Here, $\mathbf I_t^v$ denotes the visual observation from camera $v$. We learn a policy $\pi(\mathbf a_t\mid o_t,\mathbf q_t,\ell)$ to imitate demonstrated action chunks $\mathbf a_t=(a_t,\ldots,a_{t+H-1}) \in\mathbb R^{H\times d_a}$, where $H$ is the action horizon and $d_a$ is the action dimension. World-action models extend this formulation by coupling action generation with prediction of future visual states.
Let
{
\setlength{\abovedisplayskip}{3pt}
\setlength{\belowdisplayskip}{3pt}
\setlength{\abovedisplayshortskip}{0pt}
\setlength{\belowdisplayshortskip}{0pt}
\begin{align}
\mathbf o_t^+ =  (o_{t+\delta_1},\ldots,o_{t+\delta_K}), \qquad 0<\delta_1<\cdots<\delta_K,
\label{eq:future-observations}
\end{align}
}
denote $K$ future observations from the same camera views, sampled at temporal offsets $\delta_1,\ldots,\delta_K$.
Each training example $(o_t,\mathbf q_t,\ell,\mathbf o_t^+,\mathbf a_t)$ pairs the observed context with future observations and actions from the same demonstration.
In our V-JEPA Policy, the latent representations of $\mathbf o_t^+$ serve as prediction targets, and future prediction is learned jointly with action generation.
At inference, the policy is conditioned on $o_t$, $\mathbf q_t$, and $\ell$.

% codex
% In \textbf{V-JEPA Policy}, the future observations are encoded in the frozen visual latent space and used to train a future-latent predictor jointly with the action policy; the corresponding representation, information flow, and training objectives are described in Section~\ref{sec:method}.

%% file: secs/method.tex
\vspace{-1.1em}
\section{Methodology}
\label{sec:method}
\vspace{-0.8em}

\begin{figure*}[t]
\vspace{-1.3em}
\centering
\includegraphics[width=\textwidth, draft=false]{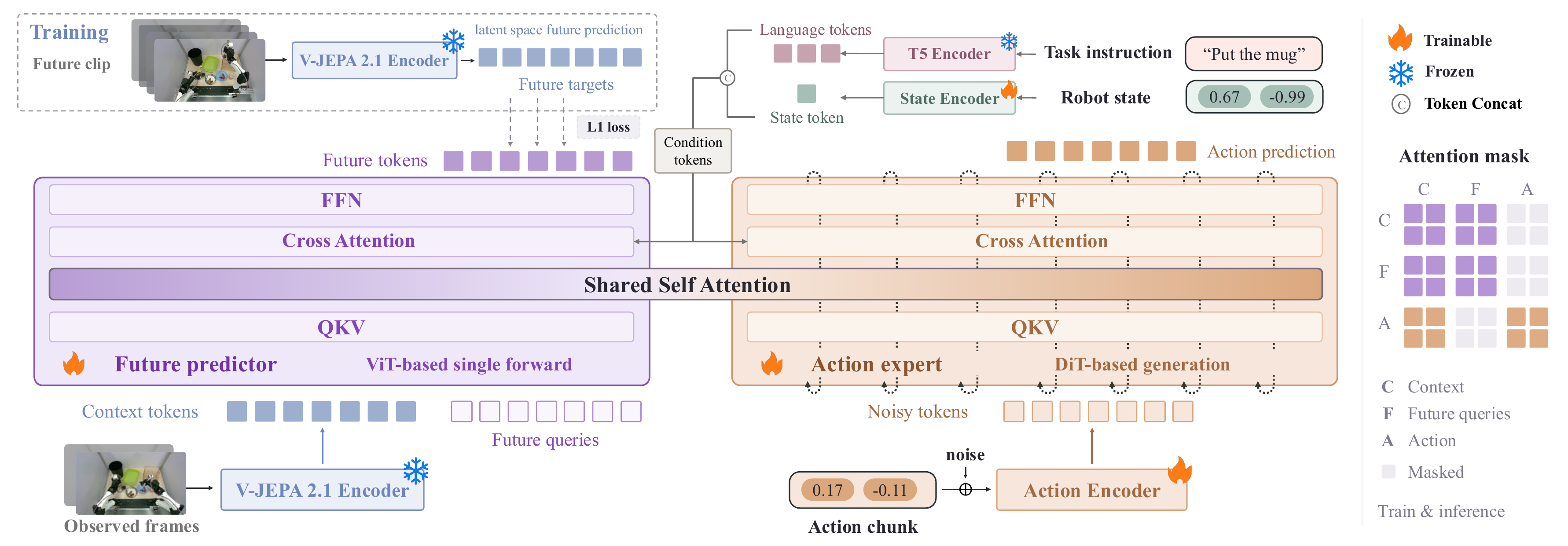}
\vspace{-2em}
% \caption{\textbf{Overview of V-JEPA Policy.}
% Built on a frozen V-JEPA 2.1 encoder, we jointly train a future-latent predictor and a flow-matching action expert from scratch, coupling future prediction with action generation through shared attention.
% }
\caption{\textbf{Overview of V-JEPA Policy.}
A frozen V-JEPA 2.1 encoder defines the visual latent space.
The future predictor jointly processes context tokens and learnable future queries in a single forward pass, providing layer-wise, future-informed context key--value states to a flow-matching action expert.
Both modules receive language and proprioceptive conditioning and are jointly trained from scratch with future-latent regression and action flow matching.
% Future observations provide training targets only.
}
\label{fig:architecture}
\vspace{-2.0em}
\end{figure*}

In this section, we first describe future-latent prediction (Sec.~\ref{sec:method-predictor}) and its coupling with action generation (Sec.~\ref{sec:method-coupling}).
We then present joint training and inference (Sec.~\ref{sec:method-training}).
Figure~\ref{fig:architecture} provides an overview of the framework.

\vspace{-1em}
\subsection{Instruction-Conditioned Latent Prediction}\label{sec:method-predictor}
\vspace{-0.6em}
\textbf{A fixed visual space.}
We use the frozen V-JEPA 2.1 encoder to encode the observed context and define the future latent targets~\citep{mur2026vjepa2_1}.
To match the encoder's temporal tubelet size of two, we form a minimal two-frame observed context for each camera view by pairing the current frame with an earlier observation.
During training, we combine this context with the corresponding future observations $\mathbf o_t^+$ to form a complete video clip.
Following the construction of context and target representations in V-JEPA~\citep{assran2025vjepa2}, we designate its future tubelet positions as the masked region.
The context path thus removes tokens at future positions before the Transformer blocks of the encoder, and encodes only observed tokens.
The target path instead processes the complete unmasked clip with the same frozen encoder. We normalize its output tokens across the feature dimension and select the future positions as prediction targets.
We independently encode each camera view and concatenate their context representations and future targets across views to form $\mathbf Z_t^c$ and $\mathbf Z_t^+$, respectively.

% For the experimental setup: We use the same two-frame observed context
% across V-JEPA encoder generations and model scales to keep visual inputs
% consistent.

\textbf{Future-query prediction.}
The predictor $P_\phi$ follows the V-JEPA 2 predictor architecture~\citep{assran2025vjepa2} and is trained from scratch to predict the future latent targets.
We project $\mathbf Z_t^c$ to the predictor's hidden dimension and concatenate the result with future-query tokens $\Delta_t^+$. These query tokens repeat a single learnable mask embedding at all future target positions.
Bidirectional self-attention jointly updates both groups, allowing context states to incorporate information from the evolving future-query representations.
To distinguish token positions, we utilize 3D rotary position embeddings (RoPE) based on temporal, height, and width coordinates in the visual latent grid~\citep{su2024rope}.
Context tokens retain the positions of the observed input, while future-query tokens use the positions of their prediction targets.
Within each attention head, we compute attention from the query, key, and value $\mathbf Q$, $\mathbf K$, and $\mathbf V$ as
% {
% \vspace{-5pt}
% \setlength{\abovedisplayskip}{3pt}
% \setlength{\belowdisplayskip}{3pt}
% \setlength{\abovedisplayshortskip}{0pt}
% \setlength{\belowdisplayshortskip}{0pt}
% \begin{equation}
% \operatorname{Attn}_{\mathrm{3D}}(\mathbf Q,\mathbf K,\mathbf V) =
% \operatorname{softmax}\!\left(
% \frac{
% \mathcal R_{\mathrm{3D}}(\mathbf Q)
% \mathcal R_{\mathrm{3D}}(\mathbf K)^{\top}
% }{
% \sqrt{d_h}
% }
% \right)\mathbf V,
% \label{eq:method-rope}
% \end{equation}
% }
{
\vspace{-4pt}
\setlength{\abovedisplayskip}{3pt}
\setlength{\belowdisplayskip}{3pt}
\setlength{\abovedisplayshortskip}{0pt}
\setlength{\belowdisplayshortskip}{0pt}
\begin{equation}
\operatorname{Attn}_{\mathrm{3D}}(\mathbf Q,\mathbf K,\mathbf V) =
\operatorname{softmax}\!\big(
\mathcal R_{\mathrm{3D}}(\mathbf Q)
\mathcal R_{\mathrm{3D}}(\mathbf K)^{\top} \!/\! \sqrt{d_h}
\big)\mathbf V,
\label{eq:method-rope}
\end{equation}
}
where $d_h$ denotes the head dimension, and $\mathcal R_{\mathrm{3D}}$ applies rotary transformations according to each token's spatiotemporal coordinates.
Coordinates are local to each camera view. Learnable view embeddings are added to both context tokens and future queries to distinguish their associated cameras.
% The visual self-attention is bidirectional over the combined sequence of context tokens and future-query tokens.
% allowing the two groups to interact throughout the predictor.

\textbf{Instruction and state conditioning.}
Future prediction is conditioned on the task instruction $\ell$ and current robot state $\mathbf q_t$.
We encode the task instruction $\ell$ with a frozen T5
encoder~\citep{raffel2019t5} and concatenate the resulting text tokens with one state token projected from $\mathbf q_t$ to form a conditioning sequence.
Following the conditioning layout of Wan~\citep{wan2025wan}, each predictor block applies visual self-attention (Eq.~\ref{eq:method-rope}), cross-attention to this sequence, and a feed-forward network.
Both context tokens and future queries therefore receive instruction and proprioceptive state information.
% The final future-query states are normalized and projected back to the encoder's latent dimension, yielding $\hat{\mathbf Z}_t^+ = P_\phi(\mathbf Z_t^c,\Delta_t^+;\ell,\mathbf q_t)$ for all future target positions in one forward pass.
The final future-query states are normalized and projected back to the encoder feature dimension, yielding the future latent predictions in a single forward pass:
{
\vspace{-1pt}
\setlength{\abovedisplayskip}{3pt}
\setlength{\belowdisplayskip}{3pt}
\setlength{\abovedisplayshortskip}{0pt}
\setlength{\belowdisplayshortskip}{0pt}
\begin{equation}
\hat{\mathbf Z}_t^+ = P_\phi\!\left( \mathbf Z_t^c,\Delta_t^+\mid\ell,\mathbf q_t \right).
\label{eq:method-future-prediction}
\end{equation}
}

% Instruction tokens from a frozen T5 encoder~\citep{raffel2019t5} are projected and concatenated with one projected state token to form a conditioning sequence.
% Inspired by the conditioning mechanism design of the diffusion transformer block in Wan~\citep{wan2025wan}, each predictor block sequentially applies bidirectional self-attention over context tokens and future queries, cross-attention to this sequence, and a feed-forward network.
% This allows context and future queries to interact while both receive task instruction and state information.
% The final future-query states are normalized and projected back to the visual latent dimension, yielding $\hat{\mathbf Z}_t^+$ for all requested positions in one forward pass.
\vspace{-1.6em}
\subsection{Coupling Prediction and Action}\label{sec:method-coupling}
\vspace{-0.6em}

%%%
\textbf{A future-informed context interface.}
We connect future prediction to action generation through the predictor's layer-wise context states rather than its final future-latent predictions.
Through bidirectional interactions with future queries in the predictor, context states at deeper layers incorporate the evolving future-query representations.
At predictor layer $j$, let $\mathbf K_c^{(j)}$ and $\mathbf V_c^{(j)}$ denote the key and value projections of the normalized block input at context positions, before its self-attention update.
The resulting layer-wise interface is $\mathcal C_\phi = \{(\mathcal R_{\mathrm{3D}}(\mathbf K_c^{(j)}), \mathbf V_c^{(j)})\}_{j=1}^{L}$, where $L$ is the number of layers.
To maintain the spatiotemporal position information of the context tubelets when conditioning the action expert, we preserve the 3D RoPE transformation applied to the keys as in Eq.~\ref{eq:method-rope} while the values remain unrotated.
This interface contains only keys and values at context positions.

\textbf{Flow-matching action generation.}
Conditioned on $\mathcal C_\phi$, the instruction $\ell$, and proprioceptive state $\mathbf q_t$, a diffusion-transformer-based action expert $v_\psi$ models continuous action chunks using conditional flow matching~\citep{peebles2023dit,liu2023flow,lipman2023flow,lipman2024flowguide}.
Given a demonstration chunk $\mathbf a_t$, Gaussian noise $\boldsymbol\epsilon\sim\mathcal N(\mathbf 0,\mathbf I)$, and flow time $\tau\in[0,1]$, we construct
{
\setlength{\abovedisplayskip}{3pt}
\setlength{\belowdisplayskip}{3pt}
\setlength{\abovedisplayshortskip}{0pt}
\setlength{\belowdisplayshortskip}{0pt}
\begin{align}
\mathbf a_t^\tau =(1-\tau)\boldsymbol\epsilon+\tau\mathbf a_t.
\label{eq:method-action-flow}
\end{align}
}
The expert embeds $\mathbf a_t^\tau$ into action tokens with learnable sequence-position embeddings.
Following recent foundation policy architectures~\citep{black2025pi05,ye2026dreamzero,zhang2026prts}, the predictor and expert form a Mixture-of-Transformers (MoT) architecture with separate parameters, the same number of layers, and compatible attention-head dimensions.
Within each attention head at layer $j$, queries projected from action tokens attend jointly to the corresponding context positions and the action tokens, and compute the output as follows:
{
\setlength{\abovedisplayskip}{3pt}
\setlength{\belowdisplayskip}{3pt}
\setlength{\abovedisplayshortskip}{0pt}
\setlength{\belowdisplayshortskip}{0pt}
\begin{equation}
\mathbf O_a^{(j)} = \operatorname{softmax}\!\big( \mathbf Q_a^{(j)} [ \mathcal R_{\mathrm{3D}}(\mathbf K_c^{(j)}); \,\mathbf K_a^{(j)} ]^{\top} \!/\! \sqrt{d_h} \big) [\mathbf V_c^{(j)};\,\mathbf V_a^{(j)}],
\label{eq:method-action-attention}
\end{equation}
}
% \begin{equation}
% \mathbf O_a^{(j)} =\operatorname{softmax}\!\Bigg(
% \frac{ \mathbf Q_a^{(j)} \left[ \mathcal R_{\mathrm{3D}}\!\left(\mathbf K_c^{(j)}\right); \,\mathbf K_a^{(j)} \right]^{\top} }{\sqrt{d_h}} \Bigg) \left[\mathbf V_c^{(j)};\,\mathbf V_a^{(j)}\right],
% \label{eq:method-action-attention}
% \end{equation}
where $\mathbf Q_a^{(j)}$, $\mathbf K_a^{(j)}$, and $\mathbf V_a^{(j)}$ are the action-token query, key, and value projections, $[\cdot;\cdot]$ denotes token concatenation, and $d_h$ is the attention-head dimension. In contrast, predictor tokens cannot attend to action tokens.
Both the predictor and action expert are conditioned on the instruction $\ell$ and proprioceptive state $\mathbf q_t$ through cross-attention.
Each module projects the T5 instruction embeddings and proprioceptive state into its own hidden space.
The output head predicts the conditional velocity $v_\psi(\mathbf a_t^\tau,\tau;\mathcal C_\phi,\ell,\mathbf q_t)$, with $\mathbf a_t-\boldsymbol\epsilon$ as the flow-matching target.
%%%

%%%%%
% 这一段还需要再改善下。
\vspace{-1.2em}
\subsection{Joint Training and Inference}\label{sec:method-training}
\vspace{-0.6em}
\textbf{Joint objective.}
For direct learning from task-specific demonstrations, we jointly train the predictor and action expert from scratch on these demonstrations in a single stage while keeping the visual and text encoders frozen.
We adapt the V-JEPA mask-denoising objective in Eq.~\ref{eq:jepa} to instruction-conditioned future prediction, with masked positions corresponding to future tubelets and targets provided by the frozen visual encoder, written as:
{
\setlength{\abovedisplayskip}{5pt}
\setlength{\belowdisplayskip}{5pt}
\setlength{\abovedisplayshortskip}{0pt}
\setlength{\belowdisplayshortskip}{0pt}
\begin{equation}
\mathcal L_{\mathrm{future}} = \mathbb{E}_{\mathcal{D}} \left\| P_\phi\!\left(\mathbf Z_t^c,\Delta_t^+\mid\ell,\mathbf q_t\right) -\mathbf Z_t^+ \right\|_1.
\label{eq:method-future-loss}
\end{equation}
}
For action generation, we optimize the flow-matching objective~\citep{lipman2023flow}
{
\setlength{\abovedisplayskip}{5pt}
\setlength{\belowdisplayskip}{5pt}
\setlength{\abovedisplayshortskip}{0pt}
\setlength{\belowdisplayshortskip}{0pt}
\begin{equation}
\mathcal L_{\mathrm{action}} = \mathbb{E}_{ ( o_t, \mathbf{o}_t^+, \mathbf{q}_t, \ell, \mathbf{a}_t ) \sim\mathcal{D}} \left\| v_\psi(\mathbf a_t^\tau,\tau;\mathcal C_\phi,\ell,\mathbf q_t) - (\mathbf a_t-\boldsymbol\epsilon)\right\|_2^2.
\label{eq:method-action-loss}
\end{equation}
}
We jointly optimize both objectives to directly learn a WAM on the frozen visual space of V-JEPA:
{
\setlength{\abovedisplayskip}{5pt}
\setlength{\belowdisplayskip}{5pt}
\setlength{\abovedisplayshortskip}{0pt}
\setlength{\belowdisplayshortskip}{0pt}
\begin{equation}
\mathcal L(\phi,\psi) = \mathcal L_{\mathrm{action}} +\lambda_{\mathrm{future}}\mathcal L_{\mathrm{future}},
\label{eq:method-joint-loss}
\end{equation}
}
with $\lambda_{\mathrm{future}}=1$.
The action loss also backpropagates through $\mathcal C_\phi$ into the predictor, so its context states are jointly shaped by the supervision of both the future prediction and the action generation.
% Predictor-only pretraining provides an alternative initialization, described in Section~\ref{sec:exp-transfer}.

\textbf{Inference.}
We follow an \emph{imagine-then-act} procedure. At each decision step, a single predictor pass constructs the future-informed context interface $\mathcal C_\phi$. We cache this interface and integrate $v_\psi$ from Gaussian noise at $\tau=0$ to a clean predicted action chunk at $\tau=1$ using 10 Euler steps, evaluating only the action expert at each step.

%% file: secs/related_work.tex
\vspace{-1.6em}
\section{Related Work}\label{sec:related-work}
\vspace{-0.8em}

\textbf{JEPA.}
Joint-embedding predictive architectures learn visual representations by predicting target embeddings from partial observations, progressing from image representation learning to video prediction and dense video features~\citep{assran2023self,bardes2024vjepa,assran2025vjepa2,mur2026vjepa2_1}.
For robot control, DINO-WM~\citep{zhou2024dinowm} and action-conditioned V-JEPA 2~\citep{assran2025vjepa2} learn dynamics in pretrained feature spaces for planning.
VLA-JEPA~\citep{vlajepa2026} integrates latent prediction with a pretrained vision-language backbone, while JEPA-WAM~\citep{lin2026jepawam} uses a Qwen-initialized predictor with an additional vision--language alignment stage before robot-policy training.
Our default framework instead jointly learns the future predictor and action expert from scratch using a frozen predictive visual encoder in a single downstream stage.
We further study how this latent space supports acquiring transferable future-modeling knowledge through predictor-only pretraining on in-the-wild action-free data.

% \textbf{World-action models.}
% World-action models couple future visual prediction with action generation, building on joint image--action prediction~\citep{wu2024unleashing} and video--action diffusion~\citep{zhu2025unifiedworldmodels}.
% A prominent line adapts pretrained video generators for control~\citep{pai2025mimicvideos,kim2026cosmos,ye2026world}, while ImageWAM uses pretrained image-editing models~\citep{zhang2026imagewam}.
% Fast-WAM retains video co-training while removing explicit future generation at inference~\citep{yuan2026fastwam}, and concurrent OpenWAM systematically compares generative backbones, latent spaces, and world--action coupling~\citep{wang2026openwam}.
% We examine predictive visual pretraining as a foundation for WAM learning through controlled encoder comparisons and the transfer of future-modeling knowledge acquired through predictor-only pretraining without action labels.

\textbf{World-action models.}
Recent WAMs acquire predictive priors by adapting pretrained video generators or image-editing models for control~\citep{liang2025videopolicy,kim2026cosmospolicy,ye2026dreamzero, yuan2026fastwam,zhang2026imagewam}.
Joint video-action modeling studies explore how visual prediction should interact with action
learning~\citep{wu2024unleashing,zhu2025unifiedworldmodels}.
FastWAM further shows that future co-training can improve policy representations even when future imagination is removed at deployment~\citep{yuan2026fastwam}.
These approaches establish several ways to use predictive knowledge while retaining pretrained generative backbones.
Our study examines the inherited visual foundation: whether the latent space learned through predictive pretraining can support effective WAM learning without inheriting a complete visual generator.

%% file: secs/experiments.tex
\vspace{-1.0em}
\section{Experiments}\label{sec:experiments}
\vspace{-0.6em}

% We extensively evaluate \textbf{V-JEPA Policy} across simulation benchmarks, distribution-shifted environments, and real-world tasks. Our experiments aim to address the following questions:
% \begin{itemize}[leftmargin=10pt, topsep=0pt,itemsep=1pt,partopsep=1pt, parsep=1pt]
%     \item Can effective WAMs be learned from task-specific demonstrations on a frozen predictive visual latent space, without inheriting a pretrained visual generator?

%     \item How does the choice of pretrained visual latent space affect performance and generalization?

%     \item Can the same latent space support acquiring future-modeling knowledge from broader in-the-wild annotated videos and transferring it to downstream control?

%     \item How does V-JEPA Policy compare with representative WAMs in training and inference cost?

% \end{itemize}

We evaluate {V-JEPA Policy} across simulation benchmarks and real-world manipulation to investigate four central questions: 
\textbf{(Q1. Viability)} Can effective WAMs be learned directly on frozen predictive visual latents without inheriting a visual generator? 
\textbf{(Q2. Visual Foundation)} How does the choice of pretrained visual latent space affect downstream performance and generalization?
\textbf{(Q3. Knowledge Transfer)} Can the same latent space support acquiring transferable future-modeling knowledge from broader video–language experience? 
\textbf{(Q4. Efficiency)} How does {V-JEPA Policy} compare with representative WAMs in inference efficiency?

\vspace{-1em}
\subsection{Experimental Setup}\label{sec:exp-setup}
\vspace{-0.5em}
\textbf{Evaluation benchmarks.} 
We evaluate V-JEPA Policy across three simulated benchmarks and a real-world dual-arm setup:
(1) \textbf{LIBERO}~\citep{liu2023libero} for in-distribution multi-task imitation across four suites (Spatial, Object, Goal, Long-Horizon);
(2) \textbf{LIBERO-Plus}~\citep{fei2025liberoplus} for out-of-distribution robustness across seven controlled shift axes (viewpoint, initial state, language, lighting, texture, noise, layout);
(3) \textbf{RoboCasa-GR1}~\citep{bjorck2025gr00t, robocasa2024} for humanoid manipulation across 24 tasks, extending evaluation to a distinct robot embodiment; and
(4) a \textbf{TianJi Marvin dual-arm platform} for real-world manipulation. 
We report success rates throughout, with evaluation protocols detailed in Appendix~\ref{sec:appendix-evaluation-protocol}.
% All benchmarks report percentage success rate (see Appendix~\ref{sec:appendix-evaluation-protocol} for details).

% % \vspace{-0.6em}
% 这里可以把下游训练预算挪到附录里面讲，在这里不讲，或者讲清楚多少个epoch即可，比如libero 保持和FastWAM一致，10个epoch
\textbf{Implementation details.}
By default, V-JEPA Policy uses a frozen V-JEPA 2.1 ViT-L visual encoder and a frozen T5-XXL text encoder.
In the default regime, we jointly train the predictor and action expert from scratch on downstream demonstrations.
We train for 10 epochs on LIBERO. On RoboCasa-GR1, we use 50k optimizer steps with a batch size
of 256.
The visual encoder, predictor, and action expert contain approximately 0.3B, 0.5B, and 0.1B parameters, respectively, totaling 0.9B parameters, of which 0.6B parameters are trainable.
Visual-foundation comparisons and future-prediction ablations in Sec.~\ref{sec:exp-foundation} use a shared downstream training recipe unless stated otherwise.
Network configurations and benchmark-specific training settings are detailed in Appendix~\ref{sec:appendix-hyperparams}.

\input{table/Q1-libero-and-plus}

% \vspace{-1em}
\subsection{Learning WAMs without a Pretrained Visual Generator}\label{sec:exp-viability}
\vspace{-0.6em}

We evaluate whether effective WAMs can be learned directly on frozen predictive visual latents by jointly training the future predictor and action expert from scratch on downstream demonstrations.
Here we also report a variant initialized with a predictor pretrained on DROID video--instruction pairs without action labels, whose pretraining protocol and transfer benefits are examined in Sec.~\ref{sec:exp-transfer}.
Neither variant uses action-supervised policy pretraining.

% We evaluate whether predictive visual latents support effective WAM learning without inheriting a pretrained visual generator.
% Therefore, we report the default model, whose predictor and action expert are jointly trained from scratch.
% More we also report a variant initialized with a predictor pretrained on DROID video--instruction pairs without action labels.
% The pretraining protocol and controlled transfer analysis are presented in Section~\ref{sec:exp-transfer}.

\textbf{Results on LIBERO and LIBERO-Plus.} Table~\ref{tab:libero-and-plus} summarizes the results on LIBERO and LIBERO-Plus.
We train on LIBERO for 10 epochs, intentionally matching the number of downstream training epochs used by FastWAM and ImageWAM. {V-JEPA Policy} achieves a 97.3\% average success rate, remaining close to the larger generator-based WAMs and competitive with representative VLAs, which introduce additional embodied policy pre-training with action labels.
This performance is obtained with only 0.9B policy parameters and no action-supervised policy pretraining.

The same policy also generalizes beyond the original LIBERO evaluation distribution.
Without additional fine-tuning on perturbed demonstrations, it reaches a 79.3\% overall success rate on LIBERO-Plus, substantially outperforming FastWAM and remaining competitive with action-pretrained VLA baselines.
ImageWAM retains a higher aggregate success rate, but the results show that competitive robustness can also be learned on frozen predictive visual latents without inheriting a complete pretrained visual generative model.

% 可以注释掉
% It should be noted that JEPA-WAM provides a closely related comparison, using an additional vision--language initialization stage before robot-policy training~\citep{lin2026jepawam}.
% Our default model achieves higher reported LIBERO success without this alignment stage, while remaining competitive under distribution shifts.
% Initializing our predictor through action-free future prediction on DROID further raises success to 98.70\% on LIBERO and 91.50\% on LIBERO-Plus, substantially exceeding JEPA-WAM.

\begin{wraptable}[7]{r}{0.55\linewidth}
\vspace{-2em}
\caption{RoboCasa-GR1 Tabletop tasks.}
\vspace{-1em}
\centering
\setlength{\tabcolsep}{2.5pt}
\footnotesize
\resizebox{\linewidth}{!}{%
\begin{tabular}{@{}lccc@{}}
\toprule
Method & Params. & Action P.T. & Success (\%) \\
\midrule
{ABot-M0}~\citep{yang2026abot}     & 4.6 & \ptYes & \textbf{58.30} \\
{GR00T}~N1.6~\citep{gr00tn1_2025}  & 3 & \ptYes & 47.60 \\
$\pi_{0.5}$~\citep{black2025pi05}         & 3.3 & \ptYes & 37.00 \\
\midrule
{StarVLA-$\pi$}~\citep{community2026starvla} & 7.8 & \ptNo  & 43.90 \\
\rowcolor{policyrow} \textbf{V-JEPA Policy} (From Scratch) & 0.9 & \ptNo  & {50.92} \\
\rowcolor{policyrow} \textbf{V-JEPA Policy} (Pretrained Predictor) & 0.9 & \ptNo  & \underline{55.58} \\
\bottomrule
\end{tabular}
}
% \vspace{-6pt}
\label{tab:robocasa}
\end{wraptable}
\textbf{Results on RoboCasa-GR1.} The same architecture and joint training objective also extend effectively to humanoid manipulation.
Across 24 RoboCasa-GR1 Tabletop tasks, {V-JEPA Policy} achieves 50.92\% success, exceeding StarVLA-$\pi$ and the action-pretrained GR00T N1.6 and $\pi_{0.5}$, while remaining below ABot-M0 (Table~\ref{tab:robocasa}).
Together with its smaller parameter count, this result supports the effectiveness of the predictive visual foundation beyond 7-DoF robotic-arm settings and demonstrates non-trivial control in high-DoF humanoid manipulation.

\begin{table*}[h]
\vspace{-0.8em}
\centering
\small
\setlength{\tabcolsep}{4pt}
\resizebox{\textwidth}{!}{%
\begin{tabular}{@{}lccccc@{}}
% \begin{tabular*}{\textwidth}{@{\extracolsep{\fill}}lccccc}
\toprule
Task & \shortstack{V-JEPA Policy \\ (From scratch)} & \shortstack{V-JEPA Policy \\ (Predictor pretrained)} & \shortstack{FastWAM \\ \citep{yuan2026fastwam}} & \shortstack{$\pi_{0.5}$ \\ \citep{black2025pi05}} & \shortstack{PRTS \\ \citep{zhang2026prts}} \\
\midrule
Table Cleanup & 11/20 (55\%) & 15/20 (75\%) & 12/20 (60\%) & \underline{17/20 (85\%)} & \textbf{20/20 (100\%)} \\
Saucer Racking & 7/20 (35\%) & \underline{17/20 (85\%)} & 7/20 (35\%) & 10/20 (50\%) & \textbf{19/20 (95\%)} \\
\bottomrule
\end{tabular}
}
\vspace{-1em}
\caption{Real-world dual-arm success rates over 20 trials per task. Task descriptions, evaluation protocols and detailed fine-tuning schedules are provided in Appendix~\ref{sec:appendix-real-world}.}
\label{tab:real-world}
\vspace{-1.4em}
\end{table*}
\textbf{Results on real-world manipulation.}
Finally, we test whether {V-JEPA Policy} remains physically feasible under real-world dynamics. 
On a four-view TianJi Marvin dual-arm platform, we evaluate five methods across two multi-stage manipulation tasks---\textit{Table Cleanup} and \textit{Saucer Racking}.
As reported in Table~\ref{tab:real-world}, the scratch policy succeeds on both physical tasks (55\% and 35\%), roughly matching generative \textsc{FastWAM} (60\% and 35\%) without requiring any visual generative pretraining. 
Initializing with an instruction-conditioned future-latent predictor pretrained on in-the-wild videos further elevates performance to 75\% and 85\%.

Considering parameter count and performance jointly, the base from-scratch policy lies on the empirical Pareto frontier of the comparison across the evaluated benchmarks and real-world tasks. We provide a visualization in Appendix~\ref{sec:appendix-pareto}.
These results show that a frozen predictive visual latent space provides a sufficient foundation for learning an effective WAM from downstream demonstrations, with the future predictor and action expert trained jointly from scratch.

% Taken together, these evaluations resolve \textbf{Q1}: learning a World-Action Model directly over frozen predictive visual latents is structurally viable and scalable. \textsc{V-JEPA Policy} matches or exceeds generative WAMs in simulation, scales to whole-body humanoid manipulation, and reliably executes physical bimanual tasks.

\vspace{-1em}
\subsection{The Visual Foundation for WAM Learning}\label{sec:exp-foundation}
\vspace{-0.5em}
To answer \textbf{Q2}, we investigate whether the choice of pretrained visual latent space influences WAM learning when the downstream prediction-control interface remains unchanged.
We therefore compare frozen visual encoders while keeping the predictor and action-expert backbones, downstream demonstrations, and optimization budget fixed.
To ensure a controlled comparison across heterogeneous visual foundations, we use the same raw video segments and align the resulting latent structures across encoders.
Each encoder receives inputs sampled according to its native patch size and temporal compression ratio, while the resulting context latents and future prediction targets are matched in spatial grid size and temporal token length.
The corresponding action chunks and labels remain identical across encoders. Encoder-specific projection layers only accommodate differences in latent feature dimensions without modifying the shared downstream WAM architecture.

\input{table/Q2-table}

\textbf{Comparing visual foundations.}
Across discriminative, reconstructive, video-understanding-oriented, and predictive representations, all evaluated encoders support downstream WAM learning, but predictive visual latents achieve the strongest overall performance under the shared recipe.
V-JEPA 2.1 ViT-L reaches 97.25\% on LIBERO and 79.25\% on LIBERO-Plus, with the advantage becoming substantially larger under distribution shifts.
Compared with the strongest discriminative baseline DINOv2, the margin increases from 2.35 points in-distribution to 12.23 points under shifts.
The comparison further reveals different generalization capabilities across representation families.
Discriminative and video-understanding-oriented representations remain competitive on appearance-related variations, reflecting their strong sensitivity to visual semantics.
In contrast, reconstructive latents exhibit substantially weaker robustness under LIBERO-Plus shifts, possibly because reconstruction objectives prioritize preserving appearance-level details that are less aligned with the object-centric and temporal abstractions required for robust control.
Overall, these results suggest that predictive visual latents provide a more compatible foundation for robust WAM learning under a unified downstream interface.

% \textbf{Comparing visual foundations.}
% Predictive visual latents yield the strongest aggregate performance among the evaluated foundations under the shared downstream recipe.
% All four V-JEPA configurations outperform the chosen discriminative, reconstructive, and video-understanding encoders on both benchmarks, with a more pronounced margin on LIBERO-Plus.
% The comparison therefore identifies the frozen latent space as an important choice for robust WAM learning, beyond selecting a high-capacity visual encoder.
% The advantage concerns overall control and generalization, rather than uniform superiority under every perturbation.

\textbf{Predictive representation quality and scale.}
Within predictive visual foundations (Table~\ref{tab:vjepa-scale}), we further examine how representation quality and encoder capacity affect downstream WAM learning.
Upgrading from V-JEPA 2 to V-JEPA 2.1 consistently improves LIBERO-Plus robustness across both ViT-L and ViT-G scales, with larger gains under distribution shifts.
In contrast, increasing encoder size from ViT-L to ViT-G primarily benefits in-distribution performance, while providing limited or even negative gains under LIBERO-Plus shifts.
Notably, V-JEPA 2.1 ViT-L achieves the strongest LIBERO-Plus performance despite using a smaller backbone than ViT-G.
These results suggest that improvements in predictive representation learning are a more reliable path toward robust WAMs than scaling visual capacity alone with fixed downstream data.

\textbf{Role of future prediction.}
We further isolate the role of explicit future-latent supervision from the contribution of the future-query pathway.
With the visual foundation and downstream recipe fixed, we compare the full model against two ablations: a context-only variant that removes future queries, and a no-future-loss variant that retains the same future-query pathway but removes future-latent supervision.
Both ablations degrade performance on LIBERO and LIBERO-Plus, with the no-future-loss variant failing to recover the full model despite preserving the future-query architecture.
This indicates that the gains do not arise merely from expanding the predictor pathway, but from explicitly learning future visual representations that are coupled to downstream action generation.
Detailed configurations and numerical results are provided in Appendix~\ref{sec:appendix-pathway}.

\vspace{-1em}
\subsection{Acquiring and Transferring Future-Modeling Knowledge}\label{sec:exp-transfer}
\vspace{-0.6em}

To investigate whether the same predictive latent space supports acquiring transferable future-modeling knowledge,
we pretrain only the instruction-conditioned predictor on DROID video--instruction pairs~\citep{khazatsky2024droid} using future-latent supervision.
The visual encoder remains frozen, and pretraining involves neither action labels nor an action expert.
We use the resulting weights to initialize the predictor for downstream joint training alongside a randomly initialized action expert.

\input{table/Q3-table}
\textbf{Transfer across benchmarks and distribution shifts.} Under matched downstream protocols, predictor-only pretraining improves success across all three simulated benchmarks and both real-world tasks (Table~\ref{tab:transfer}).
The improvement is modest on in-distribution LIBERO but substantially larger on LIBERO-Plus, where success rises from 79.25\% to 91.50\% without fine-tuning on perturbed demonstrations.
Improvements are observed across all seven perturbation axes, with the largest gains under language, texture, and robot initial state shifts (Figure~\ref{fig:transfer-combined}(b)).
Positive transfer to RoboCasa-GR1 and physical bimanual manipulation further indicates that future-modeling knowledge acquired without action supervision remains useful across robot embodiments.

\textbf{Transfer beyond downstream optimization.}
To test whether longer downstream training can recover the transfer gains, we extend a scratch run to 60k optimizer updates while keeping the model, dataset, and batch size fixed (Figure~\ref{fig:transfer-combined}(a)).
Performance improves initially but shows diminishing gains later on both LIBERO and LIBERO-Plus, remaining below the performance obtained with pretrained initialization.
On LIBERO-Plus, pretrained initialization achieves 91.50\% success rate (SR) after 10-epoch downstream updates, compared with 81.64\% after 60k updates from scratch.
Thus, additional task-specific optimization does not recover the gains from predictor pretraining along the evaluated training trajectory.

% Together, these results support a frozen predictive latent space as a reusable foundation for acquiring future-modeling knowledge from action-free video--instruction data and transferring it to downstream control.
Together, these results demonstrate that a frozen predictive latent space not only supports direct WAM learning, but also provides a reusable substrate for acquiring and transferring future-modeling knowledge from broader video--language experience.

\vspace{-1em}
\subsection{Deployment Efficiency}
\label{sec:exp-inference}
\vspace{-0.6em}
{
\setlength{\intextsep}{0pt}
\setlength{\columnsep}{8pt}
% 预留 10 行文本高度，宽度建议设为 0.50\linewidth 到 0.52\linewidth
\begin{wraptable}[6]{r}{0.62\linewidth}
\vspace{-14pt}
\centering
\footnotesize
\setlength{\tabcolsep}{2.5pt}
\caption{Action-prediction core profile on an RTX 4090.}
\label{tab:deployment-profile}
\vspace{-6pt}
\begin{tabular}{@{}lccc@{}}
\toprule
Method & Mean Latency (ms) & P95 (ms) & Peak VRAM (GiB) \\
\midrule
{PRTS}                 & \textbf{115.89} & \textbf{118.99} & 9.96 \\
$\pi_{0.5}$            & \underline{159.42} & \underline{164.46} & \underline{8.87} \\
{FastWAM}              & 202.51          & 210.72          & 12.74 \\
\rowcolor{policyrow}
\textbf{V-JEPA Policy} & 178.17          & 186.81          & \textbf{4.66} \\
\bottomrule
\end{tabular}
\vspace{-4pt}
\end{wraptable}
To address \textbf{Q4}, we compare the deployment efficiency of {V-JEPA Policy} with those of FastWAM, a representative WAM, and other baselines.
A key design choice of \textsc{V-JEPA Policy} is to perform future prediction in latent space: a single predictor forward pass produces the future-informed context interface used by the action expert, avoiding iterative visual generation at deployment.
We benchmark the action-prediction core under a matched three-view setting on an NVIDIA RTX 4090 with batch size 1, excluding sensor I/O and motor execution latency (as detailed in Appendix~\ref{sec:appendix-latency}).

As shown in Table~\ref{tab:deployment-profile}, {V-JEPA Policy} achieves a substantially smaller memory footprint among the evaluated WAMs, requiring only 4.66\,GiB peak VRAM.
Compared with FastWAM, which also avoids pixel-space decoding during deployment but retains a large visual generative backbone, {V-JEPA Policy} reduces peak memory usage by 63.4\% and
achieves lower inference latency (178.17\,ms vs.\ 202.51\,ms).
These results show that predictive latent WAMs can preserve future-conditioned action generation while avoiding the deployment overhead associated with large generative visual models.
}

%% file: table/Q1-libero-and-plus.tex
\begin{table*}[!htbp]
\vspace{-1.2em}
\centering
\small
\setlength{\tabcolsep}{2.2pt}
\renewcommand{\arraystretch}{1.08}
\resizebox{\textwidth}{!}{%
\begin{tabular}{@{}lcc|ccccc|cccccccc@{}}
% No boundary rules around the two top-level headings.
\multicolumn{3}{@{}c}{} &
\multicolumn{5}{c}{{\bf \texttt{LIBERO}}} &
\multicolumn{8}{c@{}}{{\bf \texttt{LIBERO-Plus}}} \\
\textbf{Method} &
\shortstack{\textbf{Params.}\\\textbf{(B)}} &
\shortstack{Action\\P.T.} &
% \textbf{Action \\ P.T.} &
\textbf{Avg.} & \textbf{Spatial} & \textbf{Object} &
\textbf{Goal} & \textbf{Long} &
\textbf{Avg.} & \textbf{Cam.} & \textbf{Init.} &
\textbf{Lang.} & \textbf{Light} & \textbf{Texture} &
\textbf{Noise} & \textbf{Layout} \\
\midrule
OpenVLA-OFT~\citep{kim2025fine} & 7.7 & \ptYes
& 97.1 & 97.6 & 98.4 & 97.9 & 94.5
& 69.6 & 56.4 & 31.9 & 79.5 & 88.7 & 93.3 & 75.8 & 74.2 \\
$\pi_0$~\citep{black2024pi0} & 3.3 & \ptYes
& 94.1 & 96.8 & 98.8 & 95.8 & 85.2
& 53.6 & 13.8 & 6.0 & 58.8 & 85.0 & 81.4 & 79.0 & 68.9 \\
$\pi_{0.5}$~\citep{black2025pi05} & 3.3 & \ptYes
& 96.9 & \textbf{98.8} & 98.2 & 98.0 & 92.4
& 80.7 & 64.0 & 58.0 & 88.5 & 96.6 & 81.4 & 87.5 & \underline{85.9} \\
VLA-JEPA~\citep{vlajepa2026} & 2.3 & \ptYes
& 97.2 & 96.2 & 99.6 & 97.2 & 95.8
& 79.5 & 63.3 & 67.1 & 85.4 & 95.6 & 93.6 & 66.3 & 85.1 \\
PRTS~\citep{zhang2026prts} & 5 & \ptYes
& \underline{98.4} & \textbf{98.8} & \underline{99.8} & \underline{98.4} & 96.6
& \underline{84.5} & 72.5 & 75.0 & 90.6
& 94.8 & 94.9 & 87.0 & 83.1 \\
\midrule
ResVLA~\citep{zhong2026noise} & 4 & \ptNo
& 96.6 & 96.0 & \textbf{100.0} & 97.4 & 92.8
& 76.9 & 53.2 & 57.5 & 88.2 & 94.5 & \textbf{96.3} & 81.8 & 78.3 \\
StarVLA-$\pi$~\citep{community2026starvla} & 7.8 & \ptNo
& 95.7 & \textbf{98.8} & 99.6 & 95.8 & 88.4
& 77.0 & 64.3 & 57.2 & 82.8 & 94.2 & 94.0 & 79.6 & 78.2 \\
FastWAM~\citep{yuan2026fastwam} & 6 & \ptNo
& 97.6 & \underline{98.2} & \textbf{100.0} & 97.0 & 95.2
& 51.5 & 16.4 & 44.5 & 68.9 & 78.2 & 53.7 & 37.7 & 60.7 \\
ImageWAM~\citep{zhang2026imagewam} & 4.5 & \ptNo
& \underline{98.4} & 97.2 & 99.2 & \textbf{98.8} & \underline{98.4}
& 83.1 & \textbf{80.8} & 50.3 & \underline{91.4}
& \textbf{98.1} & 85.5 & \underline{93.8} & 80.5 \\
JEPA-WAM~\citep{lin2026jepawam} & 1.3 & \ptNo
& 96.7 & 95.6 & 99.4 & {97.2} & {94.6}
& 77.9${}^{\dagger}$ & {79.2} & 59.2 & {68.2}
& {93.3} & 94.6 & {83.6} & 76.1 \\
\midrule
\rowcolor{policyrow}
\textbf{V-JEPA Policy} (From Scratch) & 0.9 & \ptNo
& 97.3 & 97.2 & 98.0 & 97.0 & 96.8
& 79.3 & 70.0 & \underline{80.7} & 64.9 & 97.2
& 80.0 & 87.8 & 79.1 \\
\rowcolor{policyrow}
\textbf{V-JEPA Policy} (Pretrained Predictor) & 0.9 & \ptNo
& \textbf{98.7} & 98.0 & 99.2 & \textbf{98.8} & \textbf{98.8}
& \textbf{91.5} & \underline{79.6} & \textbf{93.0} & \textbf{94.6} & \underline{97.4}
& \underline{95.8} & \textbf{95.3} & \textbf{87.9} \\
\bottomrule
\end{tabular}%
}
\vspace{-0.6em}
\caption{\textbf{In-distribution performance on LIBERO and generalization performance under distribution shifts on LIBERO-Plus.}
All scores are success rates (\%). Bold and underline indicate the best and second-best reported scores in each column, including ties.
{\emph{Action P.T.}} denotes additional action-supervised embodied policy pretraining.
\emph{Cam.}, \emph{Init.}, and \emph{Lang.} denote camera viewpoint, robot initial state, and language instruction perturbation axes, respectively.
${}^{\dagger}$ Micro-average estimated from the reported, rounded per-axis success rates of JEPA-WAM, weighted by the official LIBERO-Plus test-set sizes.
}
% \label{tab:libero}
\label{tab:libero-and-plus}
\vspace{-2em}
\end{table*}

%% file: table/Q2-table.tex
\begin{table}[!htbp]
\vspace{-1em}
\centering
\begingroup
% \fontfamily{ptm}\selectfont
% \fontsize{8}{10}\selectfont
\setlength{\tabcolsep}{1.6pt}
\setlength{\arrayrulewidth}{0.4pt}
\renewcommand{\arraystretch}{1.28}
\resizebox{0.98\textwidth}{!}{%
\begin{tabular}{@{}lc|ccccc|cccccccc@{}}
% \begin{tabular*}{\linewidth}{@{\extracolsep{\fill}}lc|ccccc|cccccccc@{}}
\multicolumn{2}{c|}{} & \multicolumn{5}{c|}{{\bf \texttt{LIBERO}}} & \multicolumn{8}{c}{{\bf \texttt{LIBERO-Plus}}} \\
\textbf{Visual encoder} & \textbf{Param.} & \textbf{Avg.} & \textbf{Spatial} & \textbf{Object} & \textbf{Goal} & \textbf{Long} & \textbf{Avg.} & \textbf{Cam.} & \textbf{Init.} & \textbf{Lang.} & \textbf{Light} & \textbf{Tex.} & \textbf{Noise} & \textbf{Layout} \\
\hline
\multicolumn{15}{@{}l}{\rule{0pt}{13pt}\textit{Discriminative visual foundations}} \\[1pt]
\textbf{DINOv2 ViT-L/14}~\citep{oquab2024dinov2} & 304M & 94.90 & 97.20 & 98.40 & 93.40 & 90.60 & 67.02 & 39.02 & 71.87 & 54.20 & 95.01 & 87.64 & 62.96 & 73.11 \\
\textbf{DINOv3 ViT-L/16}~\citep{simeoni2026dinov3} & 303M & 93.35 & 95.00 & 98.00 & 92.20 & 88.20 & 64.28 & 28.52 & 63.35 & 49.77 & 94.83 & 82.81 & 73.39 & 71.80 \\
\hline
\multicolumn{15}{@{}l}{\rule{0pt}{13pt}\textit{Reconstructive visual foundations}} \\[1pt]
\textbf{WAN2.2 VAE}~\citep{wan2025wan} & 150M & 92.60 & 94.40 & 97.40 & 96.20 & 82.40 & 46.43 & 35.77 & 57.29 & 45.41 & 32.05 & 10.13 & 56.53 & 73.38 \\
\hline
\multicolumn{15}{@{}l}{\rule{0pt}{13pt}\textit{Video-understanding-oriented visual foundations}} \\[1pt]
\textbf{InternVideo3}~\citep{yan2026internvideo3} & 416M & 93.10 & 94.40 & 97.80 & 96.40 & 83.80 & 68.62 & 60.73 & 72.13 & 43.79 & 88.44 & \textbf{90.71} & 64.96 & 71.80 \\
\hline
\multicolumn{15}{@{}l}{\rule{0pt}{13pt}\textit{Predictive visual foundations}} \\[1pt]
{\setlength{\fboxsep}{0.6pt}\colorbox{cyan!5}{\textbf{V-JEPA 2}}} \textbf{ViT-L}~\citep{assran2025vjepa2} & 304M & 96.90 & \underline{97.60} & 98.80 & 96.40 & \underline{94.80} & 74.15 & 52.91 & \textbf{82.77} & 59.53 & 96.58 & \underline{89.78} & 69.71 & \underline{79.21} \\
{\setlength{\fboxsep}{0.6pt}\colorbox{cyan!5}{\textbf{V-JEPA 2}}} \textbf{ViT-G}~\citep{assran2025vjepa2} & 1.01B & \textbf{97.85} & \textbf{98.60} & \underline{99.20} & \underline{97.00} & \textbf{96.80} & 76.08 & 57.85 & \underline{80.65} & 59.08 & \textbf{97.37} & 89.03 & 80.07 & 78.43 \\
\cline{1-15}
{\setlength{\fboxsep}{0.6pt}\colorbox{cyan!5}{\textbf{V-JEPA 2.1}}} \textbf{ViT-L}~\citep{mur2026vjepa2_1} & 304M & 97.25 & 97.20 & 98.00 & \underline{97.00} & \textbf{96.80} & \textbf{79.25} & \textbf{70.04} & \underline{80.65} & \textbf{64.87} & \underline{97.20} & 80.02 & \underline{87.76} & 79.08 \\
{\setlength{\fboxsep}{0.6pt}\colorbox{cyan!5}{\textbf{V-JEPA 2.1}}} \textbf{ViT-G}~\citep{mur2026vjepa2_1} & 1.01B & \underline{97.70} & \textbf{98.60} & \textbf{99.60} & \textbf{97.80} & \underline{94.80} & \underline{78.37} & \underline{61.79} & 78.52 & \underline{59.99} & 96.41 & \underline{89.78} & \textbf{89.76} & \textbf{80.66} \\
\hline
\end{tabular}
}
\endgroup
\vspace{-0.5em}
\caption{\textbf{Visual foundations, encoder generations, and model scales under a shared downstream recipe.}
All benchmark entries are success rates (\%). Bold and underline indicate the best and second-best reported scores in each column, including ties. \emph{Param.} gives the approximate frozen visual-encoder size.
% Avg. reports overall success on each benchmark; LIBERO-Plus pools test instances across its seven categories.
% Cam., Init., Lang., and Text. denote camera viewpoint, robot initial state, language, and background texture.
}
\vspace{-2.2em}
\label{tab:visual-foundations-full}
\label{tab:substrate}
\label{tab:vjepa-scale}
\end{table}

%% file: table/Q3-table.tex
\newsavebox{\qthreetransfergraphic}
\begin{figure}[t]
\vspace{-2em}
\centering
\sbox{\qthreetransfergraphic}{%
  \includegraphics[width=0.64\linewidth, draft=false]{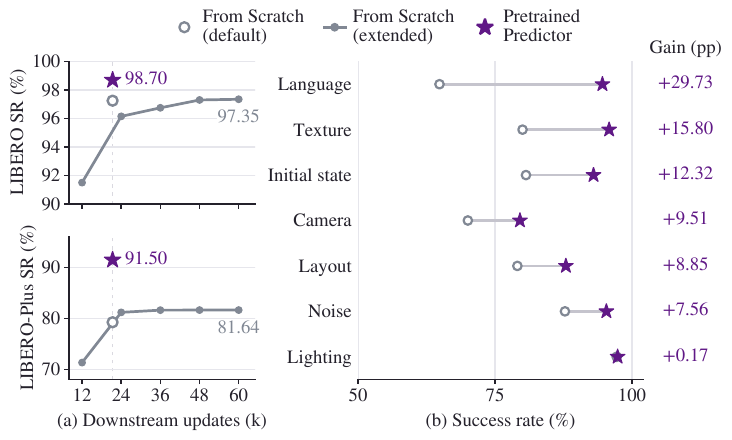}}
\begin{minipage}[t][\ht\qthreetransfergraphic][c]{0.34\linewidth}
\centering
\footnotesize
\setlength{\tabcolsep}{2pt}
\renewcommand{\arraystretch}{1.05}
\begin{tabular*}{\linewidth}{@{\extracolsep{\fill}}lcc@{}}
\toprule
Benchmark & \shortstack{\emph{From}\\\emph{Scratch}}
          & \shortstack{\emph{Pretrained}\\\emph{Predictor}} \\
\midrule
\multicolumn{3}{@{}l}{\emph{Simulation}} \\
LIBERO & 97.25 & \textbf{98.70} \\
LIBERO-Plus & 79.25 & \textbf{91.50} \\
RoboCasa-GR1 & 50.92 & \textbf{55.58} \\
\midrule
\multicolumn{3}{@{}l}{\emph{Real-world tasks}} \\
Table Cleanup & 55 & \textbf{75} \\
Saucer Racking & 35 & \textbf{85} \\
\bottomrule
\end{tabular*}
\end{minipage}\hspace{0.02\linewidth}%
\begin{minipage}[t][\ht\qthreetransfergraphic][c]{0.64\linewidth}
\centering
\usebox{\qthreetransfergraphic}
\end{minipage}
\par\nobreak\vspace{2pt}
\begin{minipage}[t]{0.34\linewidth}
\vspace{-2em}
\captionsetup{font=footnotesize,labelfont=bf,justification=justified, singlelinecheck=false,skip=0pt}
\captionof{table}{\textbf{Predictor-only pretraining improves downstream control.}
We report SR with predictors initialized from scratch or pretrained on DROID video--instruction pairs without action supervision.
Within each benchmark, only predictor initialization differs, with fixed downstream data, training protocols, and budgets.}
% \captionof{table}{\textbf{Predictor-only pretraining improves downstream control.}
% We report SR (\%) under matched downstream training protocols.
% Only predictor initialization differs: random initialization versus DROID video--instruction pretraining without action supervision. Both variants train the action expert from scratch.}
\label{tab:transfer}
\end{minipage}\hspace{0.02\linewidth}%
\begin{minipage}[t]{0.64\linewidth}
\vspace{-0.5em}
\captionsetup{font=footnotesize,labelfont=bf,justification=justified, singlelinecheck=false,skip=0pt}
% \captionof{figure}{\textbf{Transfer beyond downstream optimization.}
% (a) Curves trace a single 60k-update scratch run; separate default-scratch
% and pretrained-predictor policies use 21,360 downstream updates.
% (b) LIBERO-Plus success rates at the default budget, with pretraining gains
% in percentage points. Update counts in (a) exclude DROID pretraining.}
\captionof{figure}{\textbf{Predictor pretraining improves control and robustness beyond longer downstream training.}
(a) Success rates on LIBERO and LIBERO-Plus along an extended scratch run (\textcolor[HTML]{818894}{\emph{gray curves}}), compared with scratch (\textcolor[HTML]{818894}{\emph{open circles}}) and pretrained predictor (\textcolor[HTML]{601886}{\emph{purple stars}}) results at the default budget of 10-epoch downstream updates.
(b) Under the default budget, pretraining improves success across all seven LIBERO-Plus perturbation axes, most strongly under language shifts.
Gains are reported in percentage points (pp).}
\label{fig:transfer-combined}
\end{minipage}
\vspace{-2em}
\end{figure}

%% file: secs/appendix.tex
\section{Experimental Details}
\subsection{Evaluation Protocol}
\label{sec:appendix-evaluation-protocol}
We evaluate V-JEPA Policy on three simulation benchmarks: LIBERO, LIBERO-Plus, and RoboCasa-GR1, which assess in-distribution manipulation, robustness to controlled distribution shifts, and upper-body humanoid tabletop manipulation, respectively.

\textbf{LIBERO.}
We evaluate on all 40 tasks across the LIBERO-Spatial, LIBERO-Object, LIBERO-Goal, and LIBERO-Long suites~\citep{liu2023libero}.
For training, we use a mixed-suite dataset containing 1,693 demonstrations, obtained by excluding trajectories that fail when replayed in the simulator from the original LIBERO demonstrations~\citep{black2024pi0,black2025pi05}. 
We evaluate each task over 50 episodes and average the resulting task success rates across the ten tasks in each suite.

\textbf{LIBERO-Plus.}
LIBERO-Plus extends LIBERO with 10{,}030 perturbed tasks spanning seven categories: camera viewpoints, robot initial states, language instructions, lighting conditions, background textures, sensor noise, and object layouts~\citep{fei2025liberoplus}.
We evaluate the same policy trained on LIBERO on the full LIBERO-Plus test set, without further fine-tuning.
We report the success rate for each category and the overall success rate pooled across all tasks.

% The official category sizes reproduce 79.25 and 91.50 from the supplied
% category scores after rounding: 79.25337 and 91.49587, respectively.
% Confirm the exact evaluation manifest/version against the final run logs.

\textbf{RoboCasa-GR1.}
RoboCasa-GR1 is a simulation tabletop manipulation benchmark for the GR-1 humanoid robot, comprising 18 object-rearrangement tasks and six multi-step tasks involving articulated fixtures such as cabinets, drawers, and microwaves~\citep{bjorck2025gr00t}.
We evaluate on all 24 tasks using 50 episodes per task~\citep{yang2026abot} and report the mean success rate.

\subsection{Network Configuration and Training Recipe}\label{sec:appendix-hyperparams}

\textbf{Network configuration.}
The default model uses a frozen V-JEPA 2.1 ViT-L visual encoder and a frozen T5-XXL text encoder.
Table~\ref{tab:architecture-details} lists the predictor and action-expert configurations.
Their hidden widths differ, while their joint-attention projections have compatible head counts and dimensions.
The 4096-dimensional T5 features and proprioceptive state are projected independently into each module's hidden width.
Both modules therefore receive the instruction and current state directly.

\begin{table}[!htbp]
\centering
\small
\setlength{\tabcolsep}{5pt}
\begin{tabular*}{\linewidth}{@{\extracolsep{\fill}}lcc}
\toprule
Property & Future predictor & Action expert \\
\midrule
Transformer blocks & 24 & 24 \\
Hidden width & 1024 & 512 \\
Visual/action joint-attention heads & 16 & 16 \\
Joint-attention head dimension & 64 & 64 \\
Condition cross-attention heads & 16 & 8 \\
Condition-attention head dimension & 64 & 64 \\
Feed-forward network & GELU, expansion 4 & SwiGLU \\
Flow-time modulation & - & Adaptive RMS normalization \\
Positional encoding & 3D RoPE + view embeddings & Learned sequence embeddings \\
\bottomrule
\end{tabular*}
\caption{\textbf{Reference trainable network configuration.}
The action expert attends jointly to context-position keys and values
from the predictor and its own action tokens.
Condition cross-attention is separate from this joint attention.}
\label{tab:architecture-details}
\end{table}

\textbf{Training recipe.}
We keep the V-JEPA 2.1 visual encoder and T5-XXL text encoder frozen throughout training, and jointly optimize the future predictor and action expert on downstream tasks.
We use AdamW with a peak learning rate of $10^{-4}$, a linear warm-up over the first 5\% of scheduled steps, and subsequent cosine decay.
On LIBERO, we train jointly on the four suites for 21{,}360 optimizer steps, corresponding to ten epochs, with a global batch size of 128.
The model uses two independently encoded $224\times224$ camera views and 32-step action chunks.
On RoboCasa-GR1, we train on all 24 task datasets, each containing 1{,}000 episodes, for 50{,}000 steps with a global batch size of 256.
The model uses a single $224\times224$ egocentric view and 16-step action chunks.
For real-world experiments, we train a separate model for each of the two tasks using four camera views resized to $256\times256$.
Both tasks use 32-step action chunks and a global batch size of 64, with 60{,}000 steps for \textit{Table Cleanup} and 40{,}000 steps for \textit{Saucer Racking}.
For the pretrained-predictor variant, we first train only the future predictor on DROID for 100{,}000 steps with a global batch size of 192.
This stage uses videos, language instructions, and proprioceptive states, without action labels or an action expert.
We use two camera views resized to $256\times256$ and ten-frame clips consisting of one context tubelet and four future tubelets.
For downstream adaptation, we transfer only the pretrained predictor weights and initialize the action expert from scratch.

\section{Real-World System Implementation and Evaluation Details}
\label{sec:appendix-real-world}

\begin{wrapfigure}[14]{r}{0.38\linewidth}
\centering
\includegraphics[width=\linewidth, draft=false]{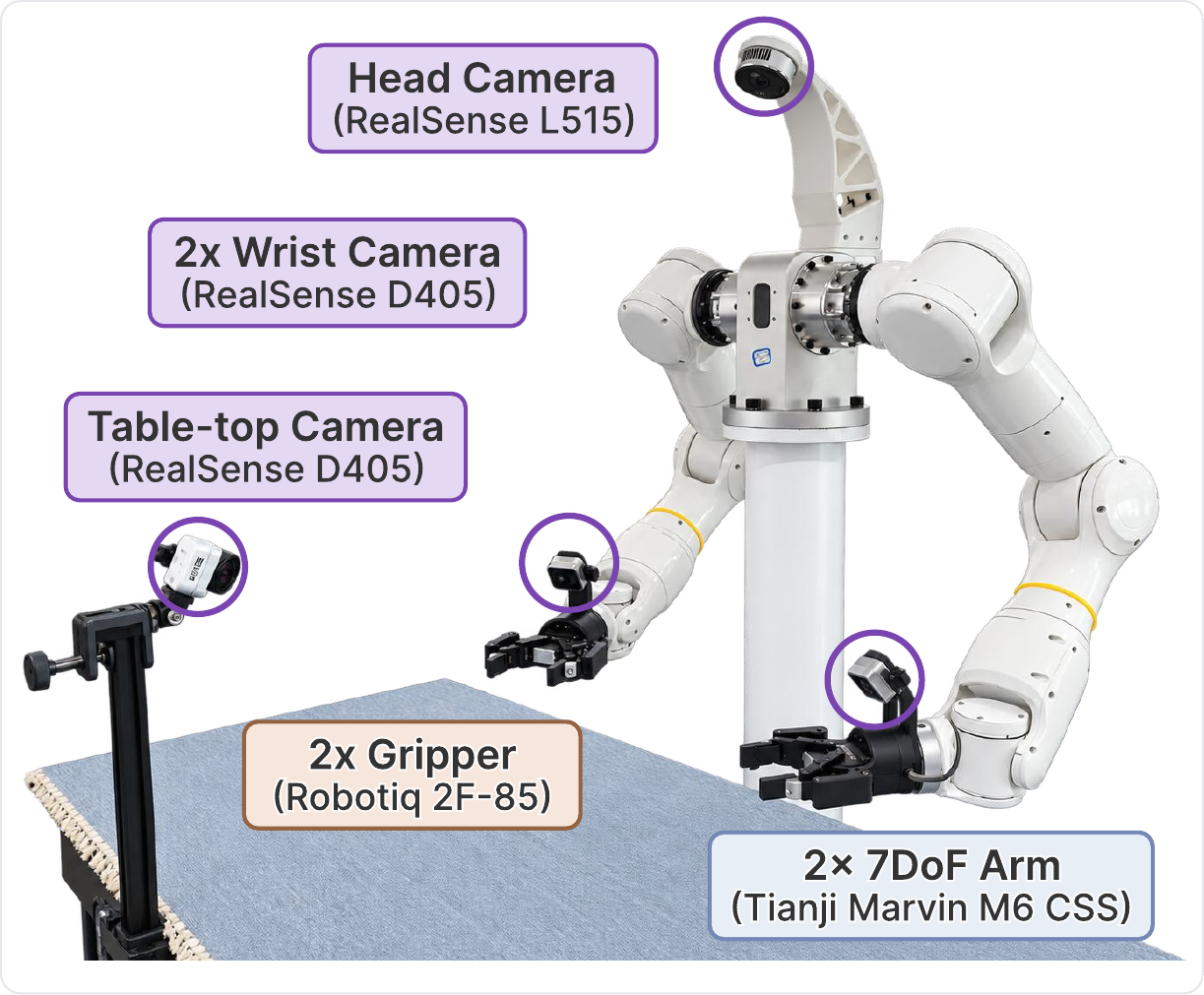}
\vspace{-2em}
\caption{TianJi Marvin dual-arm platform}
\label{fig:Tianji}
\end{wrapfigure}

\textbf{Platform and Sensor Topology.} 
Physical evaluations are deployed on a TianJi Marvin dual-arm platform comprising two 7-DoF arms, two parallel-jaw grippers, and four RGB camera streams: a head view ($1280 \times 720$) plus left wrist, right wrist, and an off-axis third-person view ($640 \times 480$ each), illustrated in Figure~\ref{fig:Tianji}. 
Policies output 16-dimensional joint-and-gripper actions (7 arm joints and 1 binary gripper state per arm).

\begin{figure}[!htbp]
\centering
\includegraphics[width=\linewidth, draft=false]{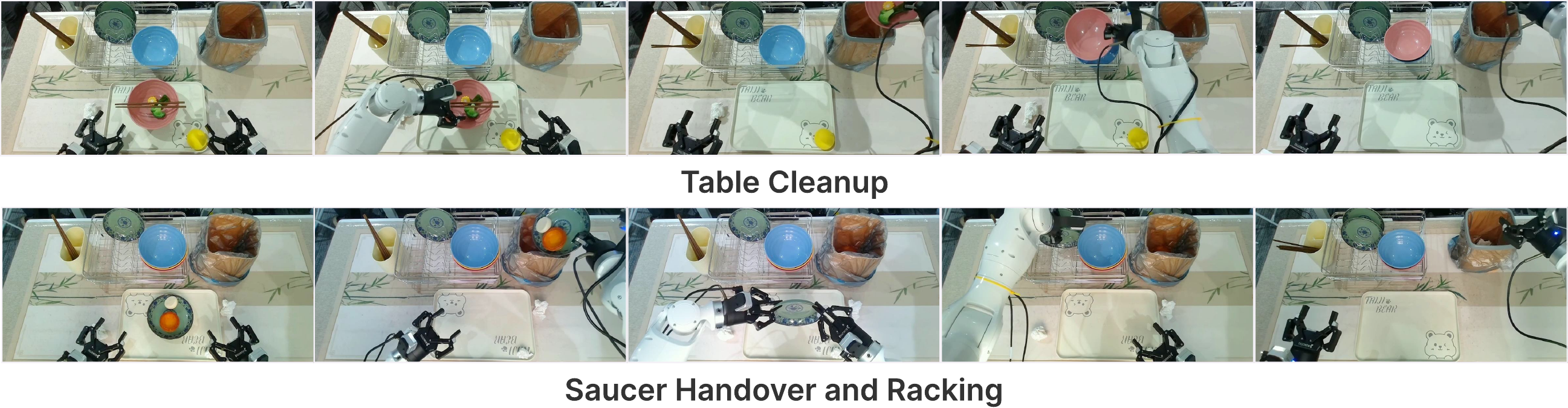}
\vspace{-0.6em}
\caption{Real-world bimanual task workflows: substage transitions and manipulation horizons.}
\label{fig:real-world-tasks}
\end{figure}

\textbf{Task Suite and Success Criteria.} 
We benchmark on two long-horizon, bimanual coordination tasks (Figure~\ref{fig:real-world-tasks}):
(1) \textit{Table Cleanup}: the left arm stows chopsticks into a holder while the right arm sweeps bowl contents into a trash bin, places the bowl in a drying rack, and discards a crumpled napkin; and 
(2) \textit{Saucer Racking}: the right arm sweeps a saucer, transfers it to the left arm via bimanual handover for rack placement, and discards residual waste. 
Trials initialize from standardized robot rest poses with objects restored to canonical configurations under invariant illumination. 
Each rollout is strictly bounded by a 120\,s execution cutoff from the first policy inference. A rollout is marked as a success only if all sequential substages are accomplished within the budget; timed-out trials are logged as 120\,s failures.

\textbf{Fine-Tuning Scope and Hyperparameters.} 
To reflect practical deployment, each method follows its native task-tuning schedule rather than an artificial compute-matched budget: \textsc{V-JEPA Policy} is fine-tuned with a batch size of 64 over 60k steps (\textit{Table Cleanup}) and 40k steps (\textit{Saucer Racking}), while baselines use a batch size of 32 over 100k and 40k steps, respectively. Consequently, real-world execution metrics benchmark practical deployed capability rather than strict sample efficiency parity.

\begin{table}[!htbp]
% \vspace{-0.4em}
\centering
\setlength{\tabcolsep}{3pt}
\begin{tabular*}{\columnwidth}{@{\extracolsep{\fill}}lcc}
\toprule
Method & Table Cleanup (s) & Saucer Racking (s) \\
\midrule
{PRTS}             & \textbf{42.40} & \textbf{42.05} \\
$\pi_{0.5}$               & \underline{54.35}    & 68.00 \\
{FastWAM}          & 62.30 & 80.86 \\
\textbf{V-JEPA Policy} (From Scratch) & 60.45 & 67.14 \\
\textbf{V-JEPA Policy} (Pretrained Predictor)   & 68.40 & \underline{59.76} \\
\bottomrule
\end{tabular*}
\caption{Mean task execution duration over successful rollouts.}
\label{tab:real-world-time}
% \vspace{-0.4em}
\end{table}

\section{Experimental Details of Visual-Foundation Comparisons}
\label{sec:appendix-encoder-comparison}
\label{sec:appendix-detailed-results}

\textbf{Comparison protocol.}
We compare frozen visual encoders using the same predictor and action-expert backbones, downstream demonstrations, global batch size of 128, and 21360 optimizer updates.
The predictor and action expert are initialized from scratch, while the visual encoder remains frozen throughout downstream training.
The evaluated foundations include discriminative features from DINOv2 and DINOv3, reconstructive latents from WAN2.2 VAE, video-understanding features from InternVideo3, and predictive features from V-JEPA 2 and V-JEPA 2.1.
Within the predictive family, we compare both encoder generations at ViT-L and ViT-G scales.

\textbf{Latent interfaces.}
Each encoder provides observed-context representations and future prediction targets in its own latent space.
Encoder-specific input and output projections accommodate differences in visual feature width while preserving the predictor's Transformer backbone.
The ViT-L variants of V-JEPA and DINO produce 1024-dimensional features, whereas WAN2.2 VAE and InternVideo3 produce 48- and 1152-dimensional features, respectively.
The action-expert architecture is unchanged across configurations.
Parameter counts in Table~\ref{tab:visual-foundations-full} include only the frozen visual encoder.

\textbf{Input alignment.}
Following Sec.~\ref{sec:exp-foundation}, we adjust spatial resolution and frame sampling within the same raw video segments to match the spatial patch grid and temporal token length across encoders, separately for observed context and future targets. For DINOv2 ViT-L/14, images are resized to $196\times196$, yielding a $14\times14$ patch grid.
This matches the grid produced by V-JEPA ViT-L with $224\times224$ inputs and $16\times16$ patches.
For WAN2.2 VAE, we subsample the input video frames with a temporal stride of two before encoding. Action labels and chunk lengths remain identical across encoder configurations.

\section{Future-Prediction Ablation}
\label{sec:appendix-pathway}

We keep the frozen visual encoder, downstream data, training seed, batch
size, update count, and action-chunk configuration fixed.
The context-only control omits future queries and conditions the action
expert on context-only predictor features.
The no-future-loss control retains the full future-query pathway and sets
the future-latent loss weight to zero.
All configurations retain action supervision.
% Future queries in the no-future-loss control can still receive gradients through the action objective.
Future queries in the no-future-loss control remain trainable and receive gradients through the action objective.

\begin{table}[!htbp]
\vspace{1em}
\centering
\small
\setlength{\tabcolsep}{3pt}
% \resizebox{\linewidth}{
\begin{tabular}{@{}ccccc@{}}
\toprule
Configuration & Future queries & Future loss & LIBERO & LIBERO-Plus \\
\midrule
Context-only & \ptNo & \ptNo & 92.55 & 65.86 \\
Future queries, no future loss & \ptYes & \ptNo & 91.65 & 68.81 \\
\rowcolor{policyrow}
\textbf{Full model} & \ptYes & \ptYes & 97.25 & 79.25 \\
\bottomrule
\end{tabular}
% }
\caption{\textbf{Future-prediction controls under the shared downstream
recipe.} All entries are success rates (\%).}
\label{tab:pathway}
\vspace{0.5em}
\end{table}

The full model outperforms both controls on both benchmarks.
The comparison with the no-future-loss control supports the contribution
of explicit future-latent supervision while keeping the query architecture
fixed.
% The context-only control changes the attention pathway, whereas the no-future-loss control changes supervision with that pathway held fixed.
The context-only control evaluates the necessity of the future-query pathway, whereas the no-future-loss control isolates the effect of explicit predictive supervision while preserving the pathway.

\section{Experimental Details of Predictor Pretraining and Transfer}
\label{sec:appendix-transfer-protocol}

\textbf{Predictor-only pretraining.}
We pretrain the instruction-conditioned predictor on DROID robot video--instruction pairs~\citep{khazatsky2024droid} using the future-latent regression objective in Eq.~\ref{eq:method-future-loss}.
Demonstrations are temporally subsampled from 15\,Hz to 5\,Hz, using one exterior camera and one wrist camera.
For each view, a training clip contains two observed frames and eight future frames.
With a temporal tubelet size of two, these correspond to one
context and four future temporal tubelet positions per view.
The predictor is additionally conditioned on the observed
proprioceptive state and cached T5 instruction features.
The visual and text encoders remain frozen, and no action expert
or action-label supervision is used during pretraining.

We use AdamW with a learning rate of $10^{-4}$, weight decay of
$10^{-2}$, and $(\beta_1,\beta_2)=(0.9,0.95)$.
Pretraining runs for 100{,}000 optimizer updates on eight GPUs
with a global batch size of 192 and BF16 precision.
The learning rate follows a 5{,}000-update linear warmup and
cosine decay over the remaining 95{,}000 updates.

\textbf{Downstream transfer.}
The pretrained predictor initializes downstream WAM learning,
while the action expert is initialized from scratch.
Both modules are then jointly optimized with the objective in
Eq.~\ref{eq:method-joint-loss}, with the visual and text
encoders kept frozen.
For each downstream benchmark or task, the scratch and
DROID-initialized models use identical demonstrations, batch sizes,
optimizer settings, and numbers of updates.
The action expert is initialized from scratch in both settings;
the predictor initialization is the only change.

On LIBERO, both models use a global batch size of 128 and 21360 optimizer updates.
The resulting checkpoints are evaluated directly on LIBERO-Plus without further fine-tuning.
On RoboCasa-GR1, both models use a batch size of 256 and 50000 optimizer updates, without gradient accumulation.
The same matched-protocol comparison is used for the real-world tasks.
Predictor pretraining adds an upstream training stage;
the matched budgets refer to downstream training, not total training computation.

\section{Action-Prediction Inference Profiling Protocol}
\label{sec:appendix-latency}

\textbf{Benchmarking Setup and Boundary Conditions.} 
Latency and memory profiles are measured on a dedicated local workstation equipped with a single NVIDIA RTX 4090 GPU (batch size 1). 
To isolate representation and model architectures from external hardware and pipeline variation, each model receives identical synthetic inputs matching our three-view setup (head, left wrist, right wrist). 
Measurements are gathered across 3 independent OS processes per architecture; each process executes 10 warmup iterations followed by 100 recorded calls, yielding 300 timed runs in total with background telemetry disabled.

Crucially, this protocol isolates the \textbf{action-prediction inference} and explicitly factors out pipeline overheads:
(1) \textit{I/O and preprocessing}: camera RTSP acquisition, image decoding, resizing, and host-to-device transfers; 
(2) \textit{Control and post-processing}: inter-process IPC, action denormalization, temporal ensembling queue updates, and One-Euro filtering; and 
(3) \textit{Actuator communication}: network packet serialization and low-level motor bus latency.

\begin{table*}[!htbp]
% \vspace{-0.4em}
\centering
\small
\setlength{\tabcolsep}{5pt}
\begin{tabular*}{\textwidth}{@{\extracolsep{\fill}}lcccccc}
\toprule
& \multicolumn{3}{c}{Action Core Latency (ms)} & \multicolumn{2}{c}{PyTorch CUDA VRAM (GiB)} & Process Memory \\
\cmidrule(lr){2-4} \cmidrule(lr){5-6} \cmidrule(lr){7-7}
Method & Mean & P50 & P95 & Allocated & Reserved & Resident (MiB) \\
\midrule
\textsc{PRTS}          & 115.89 & 115.85 & 118.99 & 9.96  & 10.30 & 11{,}014 \\
$\pi_{0.5}$            & 159.42 & 159.69 & 164.46 & 8.87  & 9.20  & 9{,}884  \\
\textsc{FastWAM}        & 202.51 & 201.93 & 210.72 & 12.74 & 13.04 & 13{,}822 \\
\textbf{V-JEPA Policy} & 178.17 & 177.72 & 186.81 & \textbf{4.66}  & \textbf{4.70}  & \textbf{5{,}276}  \\
\bottomrule
\end{tabular*}
\caption{Matched three-view action-prediction core profile on an NVIDIA RTX 4090 (300 timed runs across 3 independent processes). Allocated and Reserved denote PyTorch framework allocations; Resident denotes total host OS process memory via NVML.}
\label{tab:appendix-latency-full}
% \vspace{-0.4em}
\end{table*}

\textbf{Implementation Alignment.} 
Two adaptations ensure rigorous comparability across disparate model repositories: 
First, for {FastWAM}, generated action tensors are retained directly in GPU memory, bypassing native CPU synchronization and serialization overheads. 
Second, because the full {PRTS} and {V-JEPA Policy} checkpoints were trained on four camera streams, our benchmarked computation graphs retain the head and bimanual wrist embeddings while adapting the input sequence to three views, leaving all model weights untouched. 
Because the scratch and DROID-initialized {V-JEPA Policy} variants execute an identical computation graph, we report a single structural profile (Table~\ref{tab:appendix-latency-full}).

\section{Performance–Parameter Pareto Analysis}
\label{sec:appendix-pareto}

\begin{figure*}[!htbp]
\centering
\includegraphics[width=\textwidth, draft=false]{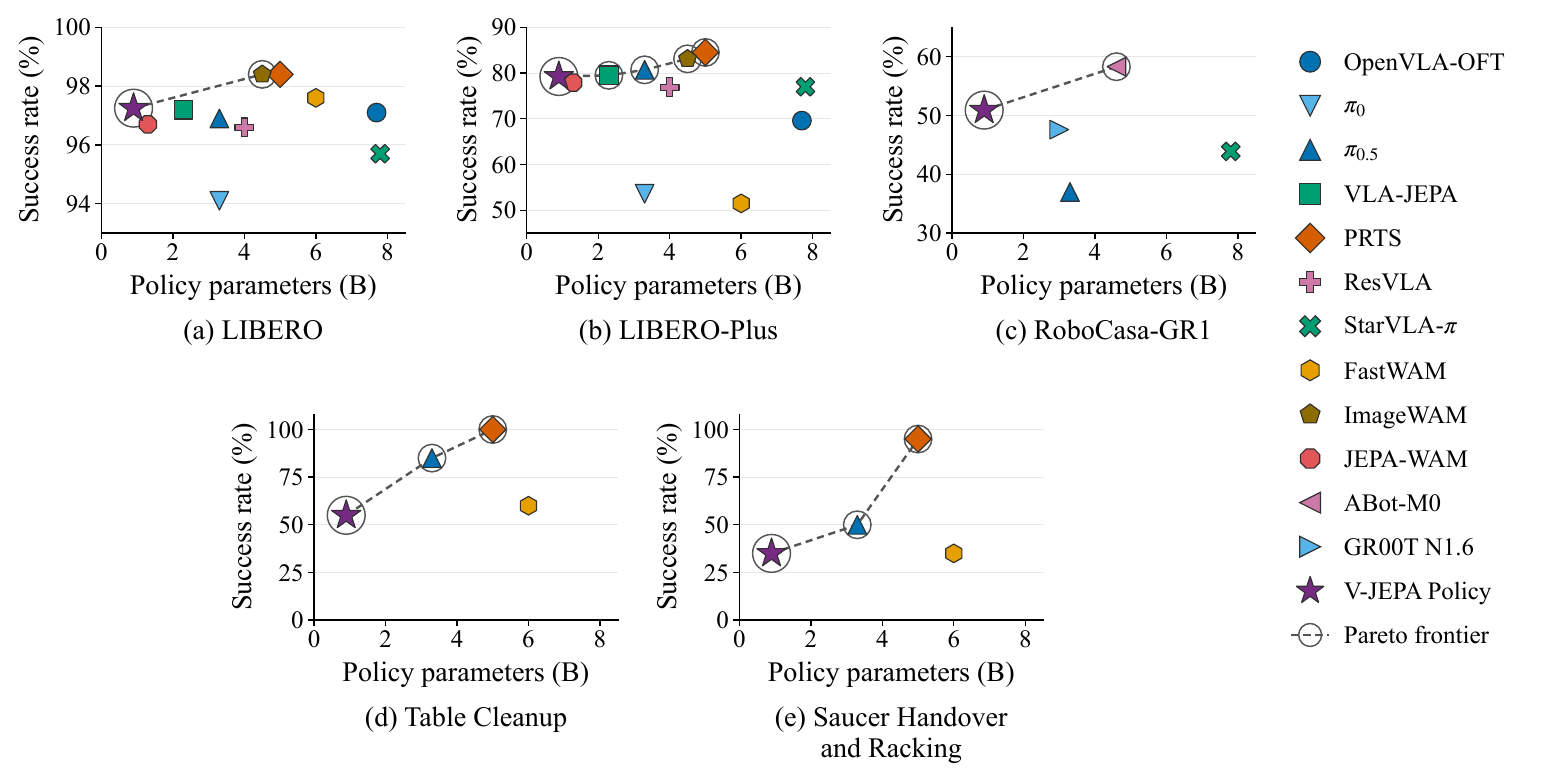}
\vspace{-1em}
\caption{Success rate versus policy parameter count across three simulation benchmarks and two real-world tasks, with circled points marking nondominated policies. The base V-JEPA Policy lies on the empirical Pareto frontier in every panel.}
\label{fig:pareto-frontier}
\vspace{-1em}
\end{figure*}

To assess the parameter efficiency of V-JEPA Policy and other baselines, we compare success rates and policy parameter counts across three simulation benchmarks and two real-world tasks. Figure~\ref{fig:pareto-frontier} presents the results. Each panel includes the baselines evaluated in that setting and the base V-JEPA Policy, whose future-latent predictor and action expert are trained from scratch on downstream demonstrations.

Each point represents a policy, with parameter count in billions on the horizontal axis and success rate on the vertical axis. 
A policy is nondominated if no other compared policy achieves at least the same success rate using no more parameters, with a strict improvement in either quantity. Circled points identify these nondominated policies, which form the empirical Pareto frontier.

The base V-JEPA Policy lies on the empirical Pareto frontier in all five settings among the compared methods. On LIBERO, it achieves 97.25\% success, 1.15 percentage points below ImageWAM’s 98.4\% while using one-fifth of its parameters. On RoboCasa-GR1, it achieves 50.92\% success with 0.9B parameters, compared with 47.60\% for the 3.0B GR00T N1.6. On the real-world tasks, it achieves 55\% success on Table Cleanup and 35\% on Saucer Racking; the latter matches FastWAM with 15\% of its parameter count. These comparisons further demonstrate that predictive visual latents provide a foundation for effective WAM learning at a compact policy scale.

\section{RoboCasa-GR1 Detailed Results}
\label{sec:robocasa-gr1-detailed}

Table~\ref{tab:robocasa-gr1-detailed} reports per-task success rates on
the 24 RoboCasa-GR1 Tabletop tasks.

\begin{table*}[t]
\centering
\caption{\textbf{Per-task success rates (\%) on RoboCasa-GR1.}
Baseline results are taken from~\citet{community2026starvla,yang2026abot},
with StarVLA-$\pi$ following its official evaluation documentation.
GR00T N1.6 task scores are rounded to integers for display;
averages are computed over all 24 tasks before rounding.
Bold indicates the best result in each row, including ties.}
\label{tab:robocasa-gr1-detailed}
\small
\setlength{\tabcolsep}{4pt}
\renewcommand{\arraystretch}{1.08}
\resizebox{\textwidth}{!}{%
\begin{tabular}{@{}lccccc@{}}
\toprule
Task
& \shortstack{ABot-M0 \\ \citep{yang2026abot}}
& \shortstack{GR00T N1.6 \\ \citep{gr00tn1_2025}}
& \shortstack{StarVLA-$\pi$ \\ \citep{community2026starvla}}
& \shortstack{V-JEPA Policy\\(From Scratch)}
& \shortstack{V-JEPA Policy\\(Pretrained Predictor)} \\
\midrule
PnPBottleToCabinetClose                    & \textbf{86} & 52 & 26 & 78 & 64 \\
PnPCanToDrawerClose                        & 74 & 13 & 62 & 82 & \textbf{86} \\
PnPCupToDrawerClose                        & \textbf{48} & 9 & 42 & \textbf{48} & \textbf{48} \\
PnPMilkToMicrowaveClose                    & 46 & 14 & 50 & 52 & \textbf{68} \\
PnPPotatoToMicrowaveClose                  & \textbf{50} & 42 & 42 & 28 & 38 \\
PnPWineToCabinetClose                      & \textbf{66} & 17 & 32 & 62 & 64 \\
\addlinespace[0.4em]
PnPNovelFromCuttingboardToBasket           & \textbf{70} & 58 & 40 & 62 & 56 \\
PnPNovelFromCuttingboardToCardboardbox     & \textbf{58} & 47 & 46 & 46 & 48 \\
PnPNovelFromCuttingboardToPan              & \textbf{76} & 69 & 60 & 62 & 70 \\
PnPNovelFromCuttingboardToPot              & \textbf{66} & 65 & 40 & 46 & \textbf{66} \\
PnPNovelFromCuttingboardToTieredbasket     & 38 & \textbf{47} & 44 & 40 & 44 \\
\addlinespace[0.4em]
PnPNovelFromPlacematToBasket               & 52 & \textbf{59} & 44 & 42 & 44 \\
PnPNovelFromPlacematToBowl                 & \textbf{66} & 58 & 52 & 48 & 44 \\
PnPNovelFromPlacematToPlate                & 60 & \textbf{63} & 50 & 58 & 58 \\
PnPNovelFromPlacematToTieredshelf          & 26 & \textbf{29} & 28 & 28 & 26 \\
\addlinespace[0.4em]
PnPNovelFromPlateToBowl                    & 54 & \textbf{57} & 52 & 46 & 56 \\
PnPNovelFromPlateToCardboardbox            & 48 & 44 & 40 & 38 & \textbf{54} \\
PnPNovelFromPlateToPan                     & \textbf{66} & 51 & 36 & 60 & 64 \\
PnPNovelFromPlateToPlate                   & 64 & \textbf{79} & 48 & 68 & 68 \\
\addlinespace[0.4em]
PnPNovelFromTrayToCardboardbox             & 54 & 52 & 34 & 54 & \textbf{68} \\
PnPNovelFromTrayToPlate                    & 68 & \textbf{71} & 64 & 64 & 58 \\
PnPNovelFromTrayToPot                      & 64 & \textbf{65} & 44 & 52 & 60 \\
PnPNovelFromTrayToTieredbasket             & \textbf{60} & 57 & 50 & 28 & 48 \\
PnPNovelFromTrayToTieredshelf              & \textbf{38} & 32 & 28 & 30 & 34 \\
\midrule
\textbf{Average}
& \textbf{58.3} & 47.6 & 43.9 & 50.92 & 55.58 \\
\bottomrule
\end{tabular}%
}
\par\vspace{0.3em}
\end{table*}

%% file: iclr2027_conference.bib
@article{ye2026dreamzero,
  title={World action models are zero-shot policies},
  author={Ye, Seonghyeon and Ge, Yunhao and Zheng, Kaiyuan and Gao, Shenyuan and Yu, Sihyun and Kurian, George and Indupuru, Suneel and Tan, You Liang and Zhu, Chuning and Xiang, Jiannan and others},
  journal={arXiv preprint arXiv:2602.15922},
  year={2026}
}

@article{yuan2026fastwam,
  title={Fast-wam: Do world action models need test-time future imagination?},
  author={Yuan, Tianyuan and Dong, Zibin and Liu, Yicheng and Zhao, Hang},
  journal={arXiv preprint arXiv:2603.16666},
  year={2026}
}

@article{liang2025videopolicy,
  title={Video generators are robot policies},
  author={Liang, Junbang and Tokmakov, Pavel and Liu, Ruoshi and Sudhakar, Sruthi and Shah, Paarth and Ambrus, Rares and Vondrick, Carl},
  journal={arXiv preprint arXiv:2508.00795},
  year={2025}
}

@inproceedings{kim2026cosmospolicy,
title={Cosmos Policy: Fine-Tuning Video Models for Visuomotor Control and Planning},
author={Moo Jin Kim and Yihuai Gao and Tsung-Yi Lin and Yen-Chen Lin and Yunhao Ge and Grace Lam and Percy Liang and Shuran Song and Ming-Yu Liu and Chelsea Finn and Jinwei Gu},
booktitle={The Fourteenth International Conference on Learning Representations},
year={2026},
url={https://openreview.net/forum?id=wPEIStHxYH}
}

@article{li2026lingbot-va,
  title={Causal world modeling for robot control},
  author={Li, Lin and Zhang, Qihang and Luo, Yiming and Yang, Shuai and Wang, Ruilin and Han, Fei and Yu, Mingrui and Gao, Zelin and Xue, Nan and Zhu, Xing and others},
  journal={arXiv preprint arXiv:2601.21998},
  year={2026}
}

@article{pai2025mimicvideos,
  title={mimic-video: Video-action models for generalizable robot control beyond vlas},
  author={Pai, Jonas and Achenbach, Liam and Montesinos, Victoriano and Forrai, Benedek and Mees, Oier and Nava, Elvis},
  journal={arXiv preprint arXiv:2512.15692},
  year={2025}
}

@article{zhang2026lingbot-va2,
  title={Native video-action pretraining for generalizable robot control},
  author={Zhang, Qihang and Li, Lin and Zhang, Luyao and Yang, Shuai and Luo, Yiming and Li, Shuaiting and Wang, Ruilin and Wang, Junke and Shao, Jiahao and Xu, Gangwei and others},
  journal={arXiv preprint arXiv:2607.08639},
  year={2026}
}

@inproceedings{bi2026motus,
  title={Motus: A unified latent action world model},
  author={Bi, Hongzhe and Tan, Hengkai and Xie, Shenghao and Wang, Zeyuan and Huang, Shuhe and Liu, Haitian and Zhao, Ruowen and Feng, Yao and Xiang, Chendong and Rong, Yinze and others},
  booktitle={Proceedings of the IEEE/CVF Conference on Computer Vision and Pattern Recognition},
  pages={35101--35113},
  year={2026}
}

@misc{bfl2025flux2,
  author    = {{Black Forest Labs}},
  title     = {{FLUX.2}: Analyzing and Enhancing the Latent Space of {FLUX} -- Representation Comparison},
  year      = {2025},
  url       = {https://bfl.ai/research/representation-comparison},
}

@article{wu2025qwen-image,
  title={Qwen-image technical report},
  author={Wu, Chenfei and Li, Jiahao and Zhou, Jingren and Lin, Junyang and Gao, Kaiyuan and Yan, Kun and Yin, Sheng-ming and Bai, Shuai and Xu, Xiao and Chen, Yilei and others},
  journal={arXiv preprint arXiv:2508.02324},
  year={2025}
}

@article{zhang2026imagewam,
  title={ImageWAM: Do World Action Models Really Need Video Generation, or Just Image Editing?},
  author={Zhang, Yuyang and Zhang, Wenyao and Qi, Zekun and Zhang, He and Lin, Haitao and Zhang, Jingbo and Mu, Yao and Yang, Xiaokang and Zeng, Wenjun and Jin, Xin},
  journal={arXiv preprint arXiv:2606.19531},
  year={2026}
}

@article{zhu2025unifiedworldmodels,
  title={Unified world models: Coupling video and action diffusion for pretraining on large robotic datasets},
  author={Zhu, Chuning and Yu, Raymond and Feng, Siyuan and Burchfiel, Benjamin and Shah, Paarth and Gupta, Abhishek},
  journal={arXiv preprint arXiv:2504.02792},
  year={2025}
}

@article{bardes2024vjepa,
  title={Revisiting feature prediction for learning visual representations from video},
  author={Bardes, Adrien and Garrido, Quentin and Ponce, Jean and Chen, Xinlei and Rabbat, Michael and LeCun, Yann and Assran, Mahmoud and Ballas, Nicolas},
  journal={arXiv preprint arXiv:2404.08471},
  year={2024}
}

@article{assran2025vjepa2,
  title={V-jepa 2: Self-supervised video models enable understanding, prediction and planning},
  author={Assran, Mido and Bardes, Adrien and Fan, David and Garrido, Quentin and Howes, Russell and Muckley, Matthew and Rizvi, Ammar and Roberts, Claire and Sinha, Koustuv and Zholus, Artem and others},
  journal={arXiv preprint arXiv:2506.09985},
  year={2025}
}

@article{mur2026vjepa2_1,
  title={V-jepa 2.1: Unlocking dense features in video self-supervised learning},
  author={Mur-Labadia, Lorenzo and Muckley, Matthew and Bar, Amir and Assran, Mido and Sinha, Koustuv and Rabbat, Mike and LeCun, Yann and Ballas, Nicolas and Bardes, Adrien},
  journal={arXiv preprint arXiv:2603.14482},
  year={2026}
}

@article{black2024pi0,
  title={$\pi_0$: A Vision-Language-Action Flow Model for General Robot Control},
  author={Black, Kevin and Brown, Noah and Driess, Danny and Esmail, Adnan and Equi, Michael and Finn, Chelsea and Fusai, Niccolo and Groom, Lachy and Hausman, Karol and Ichter, Brian and others},
  journal={arXiv preprint arXiv:2410.24164},
  year={2024}
}

@inproceedings{black2025pi05,
title={$\pi_{0.5}$: a Vision-Language-Action Model with Open-World Generalization},
author={Kevin Black and Noah Brown and James Darpinian and Karan Dhabalia and Danny Driess and Adnan Esmail and Michael Robert Equi and Chelsea Finn and Niccolo Fusai and Manuel Y. Galliker and Dibya Ghosh and Lachy Groom and Karol Hausman and brian ichter and Szymon Jakubczak and Tim Jones and Liyiming Ke and Devin LeBlanc and Sergey Levine and Adrian Li-Bell and Mohith Mothukuri and Suraj Nair and Karl Pertsch and Allen Z. Ren and Lucy Xiaoyang Shi and Laura Smith and Jost Tobias Springenberg and Kyle Stachowicz and James Tanner and Quan Vuong and Homer Walke and Anna Walling and Haohuan Wang and Lili Yu and Ury Zhilinsky},
booktitle={9th Annual Conference on Robot Learning},
year={2025},
url={https://openreview.net/forum?id=vlhoswksBO}
}

@article{zhang2026prts,
  title={PRTS: A Primitive Reasoning and Tasking System via Contrastive Representations},
  author={Zhang, Yang and Zhao, Jiangyuan and Fan, Chenyou and Yan, Fangzheng and Li, Tian and Tang, Haitong and Fu, Sen and Wu, Xuan'er and Weng, Qizhen and Zhang, Weinan and others},
  journal={arXiv preprint arXiv:2604.27472},
  year={2026}
}

@INPROCEEDINGS{khazatsky2024droid, 
    author={Khazatsky, Alexander and Pertsch, Karl and Nair, Suraj and Balakrishna, Ashwin and Dasari, Sudeep and Karamcheti, Siddharth and Nasiriany, Soroush and Srirama, Mohan Kumar and Chen, Lawrence Yunliang and Ellis, Kirsty and others},
    TITLE     = {{DROID: A Large-Scale In-The-Wild Robot Manipulation Dataset}}, 
    BOOKTITLE = {Proceedings of Robotics: Science and Systems}, 
    YEAR      = {2024}, 
    ADDRESS   = {Delft, Netherlands}, 
    MONTH     = {July}, 
    DOI       = {10.15607/RSS.2024.XX.120} 
}

@article{oquab2024dinov2,
title={{DINO}v2: Learning Robust Visual Features without Supervision},
author={Maxime Oquab and Timoth{\'e}e Darcet and Th{\'e}o Moutakanni and Huy V. Vo and Marc Szafraniec and Vasil Khalidov and Pierre Fernandez and Daniel HAZIZA and Francisco Massa and Alaaeldin El-Nouby and Mido Assran and Nicolas Ballas and Wojciech Galuba and Russell Howes and Po-Yao Huang and Shang-Wen Li and Ishan Misra and Michael Rabbat and Vasu Sharma and Gabriel Synnaeve and Hu Xu and Herve Jegou and Julien Mairal and Patrick Labatut and Armand Joulin and Piotr Bojanowski},
journal={Transactions on Machine Learning Research},
issn={2835-8856},
year={2024},
url={https://openreview.net/forum?id=a68SUt6zFt},
note={Featured Certification}
}

@article{simeoni2026dinov3,
title={{DINO}v3},
author={Oriane Sim{\'e}oni and Huy V. Vo and Maximilian Seitzer and Federico Baldassarre and Maxime Oquab and Cijo Jose and Vasil Khalidov and Marc Szafraniec and Seung Eun Yi and Michael Ramamonjisoa and Francisco Massa and Daniel HAZIZA and Luca Wehrstedt and Jianyuan Wang and Timoth{\'e}e Darcet and Th{\'e}o Moutakanni and Leonel Sentana and Claire Roberts and Andrea Vedaldi and Jamie Tolan and John Brandt and Camille Couprie and Julien Mairal and Herve Jegou and Patrick Labatut and Piotr Bojanowski},
journal={Transactions on Machine Learning Research},
issn={2835-8856},
year={2026},
url={https://openreview.net/forum?id=2NlGyqNjns},
note={Featured Certification}
}

@article{wan2025wan,
  title={Wan: Open and advanced large-scale video generative models},
  author={Wan, Team and Wang, Ang and Ai, Baole and Wen, Bin and Mao, Chaojie and Xie, Chen-Wei and Chen, Di and Yu, Feiwu and Zhao, Haiming and Yang, Jianxiao and others},
  journal={arXiv preprint arXiv:2503.20314},
  year={2025}
}

@article{yan2026internvideo3,
  title={InternVideo3: Agentify Foundation Models with Multimodal Contextual Reasoning},
  author={Yan, Ziang and Xia, Sheng and Yu, Jiashuo and Wu, Yue and Jiang, Tianxiang and Li, Songze and Tian, Kanghui and Xu, Yicheng and He, Yinan and Chen, Kai and others},
  journal={arXiv preprint arXiv:2606.12195},
  year={2026}
}

@article{liu2023libero,
  title={LIBERO: Benchmarking Knowledge Transfer for Lifelong Robot Learning},
  author={Liu, Bo and Zhu, Yifeng and Gao, Chongkai and Feng, Yihao and Liu, Qiang and Zhu, Yuke and Stone, Peter},
  journal={arXiv preprint arXiv:2306.03310},
  year={2023}
}

@article{fei2025liberoplus,
  title={Libero-plus: In-depth robustness analysis of vision-language-action models},
  author={Fei, Senyu and Wang, Siyin and Shi, Junhao and Dai, Zihao and Cai, Jikun and Qian, Pengfang and Ji, Li and He, Xinzhe and Zhang, Shiduo and Fei, Zhaoye and others},
  journal={arXiv preprint arXiv:2510.13626},
  year={2025}
}

@inproceedings{robocasa2024,
  title={RoboCasa: Large-Scale Simulation of Everyday Tasks for Generalist Robots},
  author={Soroush Nasiriany and Abhiram Maddukuri and Lance Zhang and Adeet Parikh and Aaron Lo and Abhishek Joshi and Ajay Mandlekar and Yuke Zhu},
  booktitle={Robotics: Science and Systems (RSS)},
  year={2024}
}

@article{bjorck2025gr00t,
  title={Gr00t n1: An open foundation model for generalist humanoid robots},
  author={Bjorck, Johan and Casta{\~n}eda, Fernando and Cherniadev, Nikita and Da, Xingye and Ding, Runyu and Fan, Linxi and Fang, Yu and Fox, Dieter and Hu, Fengyuan and Huang, Spencer and others},
  journal={arXiv preprint arXiv:2503.14734},
  year={2025}
}

@article{su2024rope,
  title={Roformer: Enhanced transformer with rotary position embedding},
  author={Su, Jianlin and Ahmed, Murtadha and Lu, Yu and Pan, Shengfeng and Bo, Wen and Liu, Yunfeng},
  journal={Neurocomputing},
  volume={568},
  pages={127063},
  year={2024},
  publisher={Elsevier}
}

@article{lecun2022path,
  title={A path towards autonomous machine intelligence version 0.9. 2, 2022-06-27},
  author={LeCun, Yann and others},
  journal={Open Review},
  volume={62},
  number={1},
  pages={1--62},
  year={2022}
}

@article{dosovitskiy2020vit,
  title={An image is worth 16x16 words: Transformers for image recognition at scale},
  author={Dosovitskiy, Alexey and Beyer, Lucas and Kolesnikov, Alexander and Weissenborn, Dirk and Zhai, Xiaohua and Unterthiner, Thomas and Dehghani, Mostafa and Minderer, Matthias and Heigold, Georg and Gelly, Sylvain and others},
  journal={arXiv preprint arXiv:2010.11929},
  year={2020}
}

@misc{vlajepa2026,
  title={VLA-JEPA: Enhancing Vision-Language-Action Model with Latent World Model}, 
  author={Jingwen Sun and Wenyao Zhang and Zekun Qi and Shaojie Ren and Zezhi Liu and Hanxin Zhu and Guangzhong Sun and Xin Jin and Zhibo Chen},
  year={2026},
  eprint={2602.10098},
  archivePrefix={arXiv},
  primaryClass={cs.RO},
  url={https://arxiv.org/abs/2602.10098}, 
}

@article{kim2025fine,
    title={Fine-Tuning Vision-Language-Action Models: Optimizing Speed and Success},
    author={Kim, Moo Jin and Finn, Chelsea and Liang, Percy},
    journal={arXiv preprint arXiv:2502.19645},
    year={2025}
}

@article{community2026starvla,
    title={StarVLA: A Lego-like Codebase for Vision-Language-Action Model Developing},
    author={Community, StarVLA},
    journal={arXiv preprint arXiv:2604.05014},
    year={2026},
    eprint={2604.05014},
    archivePrefix={arXiv},
    primaryClass={cs.RO}
}

@article{zhong2026noise,
  title={From Noise to Intent: Anchoring Generative VLA Policies with Residual Bridges},
  author={Zhong, Yiming and He, Yaoyu and Yang, Zemin and Tian, Pengfei and Huang, Yifan and Huang, Qingqiu and Zhu, Xinge and Ma, Yuexin},
  journal={arXiv preprint arXiv:2604.21391},
  year={2026}
}

@article{yang2026abot,
  title={ABot-M0: VLA Foundation Model for Robotic Manipulation with Action Manifold Learning},
  author={Yang, Yandan and Zeng, Shuang and Lin, Tong and Chang, Xinyuan and Qi, Dekang and Xiao, Junjin and Liu, Haoyun and Chen, Ronghan and Chen, Yuzhi and Huo, Dongjie and others},
  journal={arXiv preprint arXiv:2602.11236},
  year={2026}
}

@inproceedings{gr00tn1_2025,
  archivePrefix = {arxiv},
  eprint     = {2503.14734},
  title      = {{GR00T} {N1}: An Open Foundation Model for Generalist Humanoid Robots},
  author     = {NVIDIA and Johan Bjorck and Fernando Castañeda, Nikita Cherniadev and Xingye Da and Runyu Ding and Linxi "Jim" Fan and Yu Fang and Dieter Fox and Fengyuan Hu and Spencer Huang and Joel Jang and Zhenyu Jiang and Jan Kautz and Kaushil Kundalia and Lawrence Lao and Zhiqi Li and Zongyu Lin and Kevin Lin and Guilin Liu and Edith Llontop and Loic Magne and Ajay Mandlekar and Avnish Narayan and Soroush Nasiriany and Scott Reed and You Liang Tan and Guanzhi Wang and Zu Wang and Jing Wang and Qi Wang and Jiannan Xiang and Yuqi Xie and Yinzhen Xu and Zhenjia Xu and Seonghyeon Ye and Zhiding Yu and Ao Zhang and Hao Zhang and Yizhou Zhao and Ruijie Zheng and Yuke Zhu},
  month      = {March},
  year       = {2025},
  booktitle  = {ArXiv Preprint},
}

@InProceedings{peebles2023dit,
    author    = {Peebles, William and Xie, Saining},
    title     = {Scalable Diffusion Models with Transformers},
    booktitle = {Proceedings of the IEEE/CVF International Conference on Computer Vision (ICCV)},
    month     = {October},
    year      = {2023},
    pages     = {4195-4205}
}

@inproceedings{liu2023flow,
title={Flow Straight and Fast: Learning to Generate and Transfer Data with Rectified Flow},
author={Xingchao Liu and Chengyue Gong and qiang liu},
booktitle={The Eleventh International Conference on Learning Representations },
year={2023},
url={https://openreview.net/forum?id=XVjTT1nw5z}
}

@inproceedings{lipman2023flow,
    title={Flow Matching for Generative Modeling},
    author={Yaron Lipman and Ricky T. Q. Chen and Heli Ben-Hamu and Maximilian Nickel and Matthew Le},
    booktitle={The Eleventh International Conference on Learning Representations },
    year={2023},
    url={https://openreview.net/forum?id=PqvMRDCJT9t}
}

@article{lipman2024flowguide,
  title={Flow matching guide and code},
  author={Lipman, Yaron and Havasi, Marton and Holderrieth, Peter and Shaul, Neta and Le, Matt and Karrer, Brian and Chen, Ricky TQ and Lopez-Paz, David and Ben-Hamu, Heli and Gat, Itai},
  journal={arXiv preprint arXiv:2412.06264},
  year={2024}
}

@article{raffel2019t5,
  title={Exploring the limits of transfer learning with a unified text-to-text transformer},
  author={Raffel, Colin and Shazeer, Noam and Roberts, Adam and Lee, Katherine and Narang, Sharan and Matena, Michael and Zhou, Yanqi and Li, Wei and Liu, Peter J},
  journal={Journal of machine learning research},
  volume={21},
  number={140},
  pages={1--67},
  year={2020}
}

@inproceedings{assran2023self,
  title={Self-supervised learning from images with a joint-embedding predictive architecture},
  author={Assran, Mahmoud and Duval, Quentin and Misra, Ishan and Bojanowski, Piotr and Vincent, Pascal and Rabbat, Michael and LeCun, Yann and Ballas, Nicolas},
  booktitle={2023 IEEE/CVF Conference on Computer Vision and Pattern Recognition (CVPR)},
  pages={15619--15629},
  year={2023},
  organization={IEEE}
}

@article{zhou2024dinowm,
  title={Dino-wm: World models on pre-trained visual features enable zero-shot planning},
  author={Zhou, Gaoyue and Pan, Hengkai and LeCun, Yann and Pinto, Lerrel},
  journal={arXiv preprint arXiv:2411.04983},
  year={2024}
}

@article{lin2026jepawam,
  title={JEPA-WAM: Learning vision-language-action policies with joint-embedding world modeling},
  author={Lin, Yihan and He, Jiawei and Bao, Shifeng and Zhao, Chen and Li, Yang and Wang, Xiaobo and Wang, Yan and Chi, Cheng and Zhang, Jing},
  journal={arXiv preprint arXiv:2608.09381},
  year={2026}
}

@inproceedings{wu2024unleashing,
title={Unleashing Large-Scale Video Generative Pre-training for Visual Robot Manipulation},
author={Hongtao Wu and Ya Jing and Chilam Cheang and Guangzeng Chen and Jiafeng Xu and Xinghang Li and Minghuan Liu and Hang Li and Tao Kong},
booktitle={The Twelfth International Conference on Learning Representations},
year={2024},
url={https://openreview.net/forum?id=NxoFmGgWC9}
}
